\pdfoutput=1

\documentclass[11pt]{article}

\usepackage[preprint]{acl}

\usepackage{times}
\usepackage{latexsym}

\usepackage[T1]{fontenc}

\usepackage[utf8]{inputenc}

\usepackage{microtype}
\usepackage{makecell}

\usepackage{inconsolata}

\usepackage{graphicx}

\usepackage{subfigure}

\usepackage{breakurl} 

\usepackage{booktabs} 

\usepackage{tabularx}

\usepackage{graphicx} 

\usepackage{adjustbox}

\usepackage{float}

\newcommand*\rot{\rotatebox{90}}

\title{Unveiling Dual Quality in Product Reviews: An NLP-Based Approach}

\author{Rafał Poświata, Marcin Michał Mirończuk, Sławomir Dadas, \\ \bf Małgorzata Grębowiec, Michał Perełkiewicz\\
  National Information Processing Institute \\
  al. Niepodległości 188b, 00-608 Warsaw, Poland \\
  \texttt{\{rposwiata, mmironczuk, sdadas, mgrebowiec, mperelkiewicz\}@opi.org.pl} \\}

\begin{document}
\maketitle

\begin{abstract}
Consumers often face inconsistent product quality, particularly when identical products vary between markets, a situation known as the dual quality problem. To identify and address this issue, automated techniques are needed. This paper explores how natural language processing (NLP) can aid in detecting such discrepancies and presents the full process of developing a solution.
First, we describe in detail the creation of a new Polish-language dataset with 1,957 reviews, 540 highlighting dual quality issues. We then discuss experiments with various approaches like SetFit with sentence-transformers, transformer-based encoders, and LLMs, including error analysis and robustness verification. Additionally, we evaluate multilingual transfer using a subset of opinions in English, French, and German. The paper concludes with insights on deployment and practical applications.
\end{abstract}

\section{Introduction}

Dual quality of products refers to practices where companies sell items under the same brand and similar packaging in different markets, yet present them with significantly altered composition or quality parameters~\cite{2018BEUC}. This phenomenon has sparked growing controversy among consumers, especially within the European Union (EU), where it is perceived as a potential violation of fair competition rules~\cite{2018BEUC}. From a sociological and economic perspective, dual quality practices raise multifaceted concerns about market trust, purchasing behaviours and the perception of fairness among consumers~\cite{2021lenka,2021blucia}. Multiple reports published by consumer organizations and EU research services suggest that offering products with distinct ingredients or characteristics under identical branding constitutes a widespread international issue~\cite{2018BEUC,2019EPRS,2023ECJRC}. 
The above reasons and EU regulations—such as the amended Directive on Unfair Commercial Practices—recognize dual quality as misleading conduct, which may require enforcement at the national level \cite{2025Chambers,2025EUMonitor} (also, see more details in Appendix~\ref{appx:dq_regulations}). Our recent research project focused on creating a solution to support a national agency from one of the EU countries to address the above problem, namely the Office of Competition and Consumer Protection (UOKiK) in Poland ({\url{https://uokik.gov.pl/en}). 

The main goal of the project was to automate the detection of unfair commercial practices using natural language processing (NLP) methods. The project, currently in the proof-of-concept stage, is enabling the automated collection and analysis of product-related data from e-commerce sites and social media. It comprises a data retrieval module (intelligent web crawling, scraping, cleaning, and preprocessing) and a text analysis module that includes language identification, sentiment analysis, aspect base sentiment analysis, and the detection of consumer reviews\footnote{In this article, we use the terms ‘reviews’ and ‘opinions’ interchangeably to refer to consumer expressions regarding a product. While ‘review’ may often imply a structured evaluation, we also include informal opinions that may indicate perceptions of dual quality.} that may indicate potential dual quality issues in products.

In this paper, we focus on the last and most novel of these components for detecting dual quality reviews, describing the entire process from data preparation, through extensive evaluation of different approaches, to deployment. To our knowledge, no available dataset or model is aimed at recognizing dual quality-related reviews. While several articles (discussed further in Section~\ref{sec:relwork}) approach dual quality from sociological, economic, and legal perspectives, our study takes a different approach presented in Figure~\ref{fig:dqrec}. 

\begin{figure}[!htb]
\centering
\includegraphics[scale=0.38]{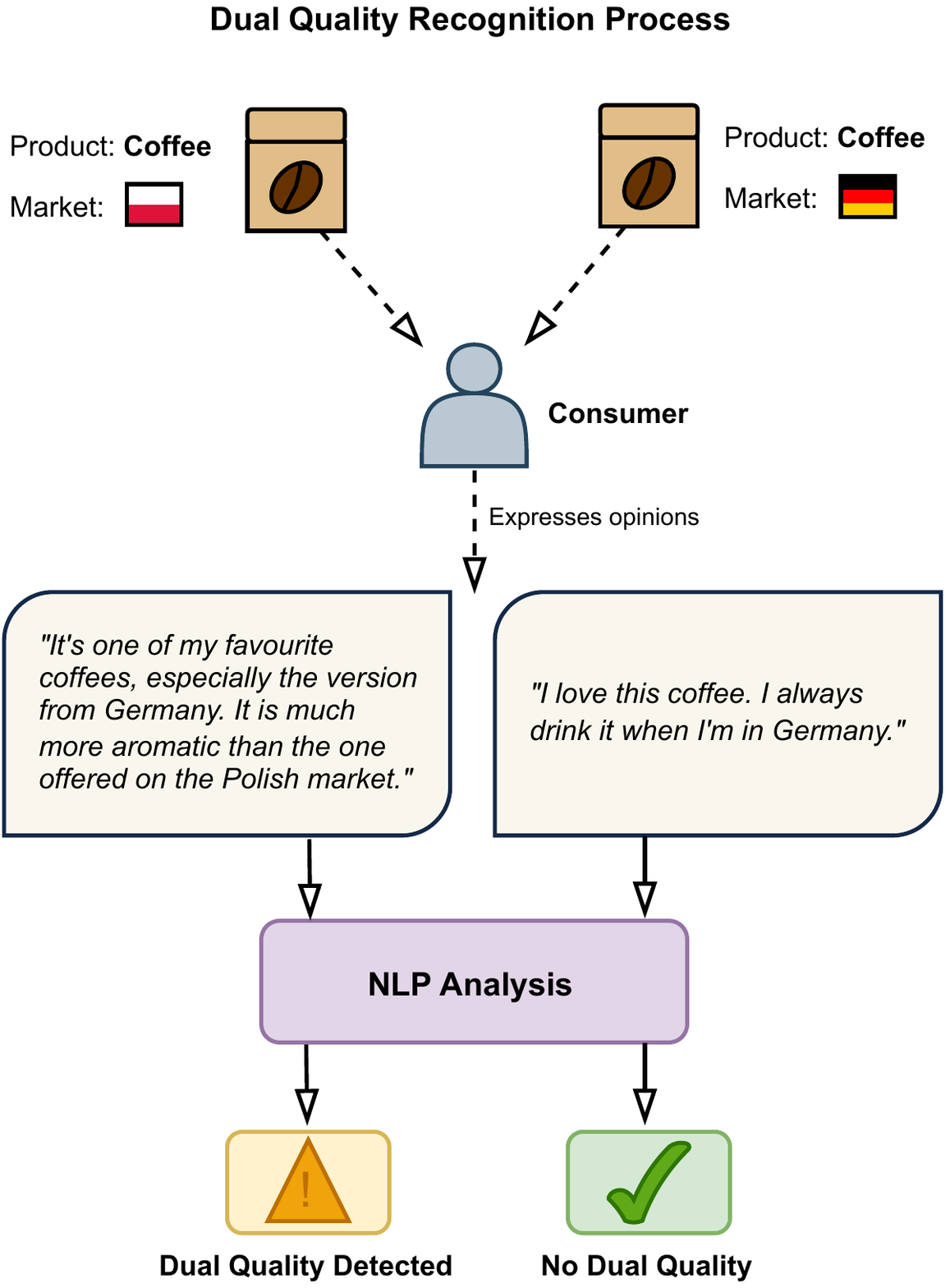}
\caption{Illustration of the NLP-based workflow for recognizing dual quality consumer reviews. The dual quality detection system flags reviews for potential issues when a consumer explicitly notes a difference between product versions from different markets. This illustration exemplifies the process with a Polish consumer assessing products from Polish and German markets; the reviews shown are English translations of the original Polish texts for clarity and wider accessibility.}
\label{fig:dqrec}
\end{figure}

The main contributions of this work can be summarized as follows:

\noindent
\textbf{--} Proposition of new NLP task: detecting the dual quality issues in product reviews. 

\noindent
\textbf{--} A coherent methodology for dataset construction and preparation of a corpus of 1,957 human-verified product reviews, 540 of which potentially exhibit dual quality. 

\noindent
\textbf{--} A comprehensive evaluation of Polish and multilingual models, including a presentation of various metrics, error analysis, and robustness verification conducted primarily for Polish.

\noindent
\textbf{--} Expansion of the dataset to include product reviews in other key languages such as English, German, and French, demonstrating the system's multilingual capabilities.}

\section{Related Work}\label{sec:relwork}

Economic and social research on dual quality products highlights the erosion of consumer trust when identical branding masks disparities in product quality across EU Member States. Studies indicate that these discrepancies, particularly in food products, impact consumer perceptions of fairness and lead to behavioral changes in purchasing decisions~\citep{2018lucia, 2019lucia, 2021lucia, 2021blucia}. Research has further demonstrated that wealthier consumers are more aware of the issue and seek alternatives in other markets, whereas lower-income consumers are more likely to adapt their behavior to avoid lower-quality products~\citep{2021blucia}. The perception of dual quality as an economic problem is also evident, as lower-quality ingredients often correspond to price disparities that disadvantage consumers in specific regions~\citep{2020janzav}.

Additionally, empirical studies confirm that public perception of dual quality is shaped by exposure to media reports and political discourse, leading to heightened scrutiny of multinational corporations and their regional product differentiation strategies~\citep{2021lenka}. While some scholars argue that manufacturers may justify product variations based on local market preferences, research suggests that these practices often lack transparency and leave consumers feeling deceived~\citep{2023lenka}. Moreover, comparative consumer tests confirm that dual quality is not confined to food products but also extends to household and personal care items, reinforcing the need for regulatory intervention~\citep{2024lucia}. Given the strong consumer opposition across Europe, particularly in Central and Eastern European countries, economic research increasingly supports regulatory measures to curb these practices and ensure consistent product quality across EU markets.

From an computer science perspective, the topic of applying NLP techniques to e-commerce platforms and customer behavior analysis is widely studied. Among these works, we can point out customer reviews analysis~\citep{ivebotunax2024, joniuhansa2024, yonatanmam2024}, product question answering~\cite{shen2023, wang2023}, product categorization~\cite{gong2023}, moderation of e-commerce reviews~\cite{nayak2022}, product feature extraction from the web~\cite{fuchs2022}, customer service support~\cite{obadinma2022}, data augmentation in e-commerce~\cite{avigdor2023}, fake news detection~\cite{hu2023}, predictive quality in manufacturing~\cite{2022tercan}, or intent classification~\cite{parikh2023}. However, none of these works address the dual quality problem directly or consider how to harness consumer opinions—such as reviews from the Internet, e-commerce platforms, or social media—to help resolve this issue. Thus, a clear research gap exists in applying NLP-based methods to detect or analyze dual quality products.

\begin{figure*}[!ht]
\centering
\includegraphics[scale=0.455]{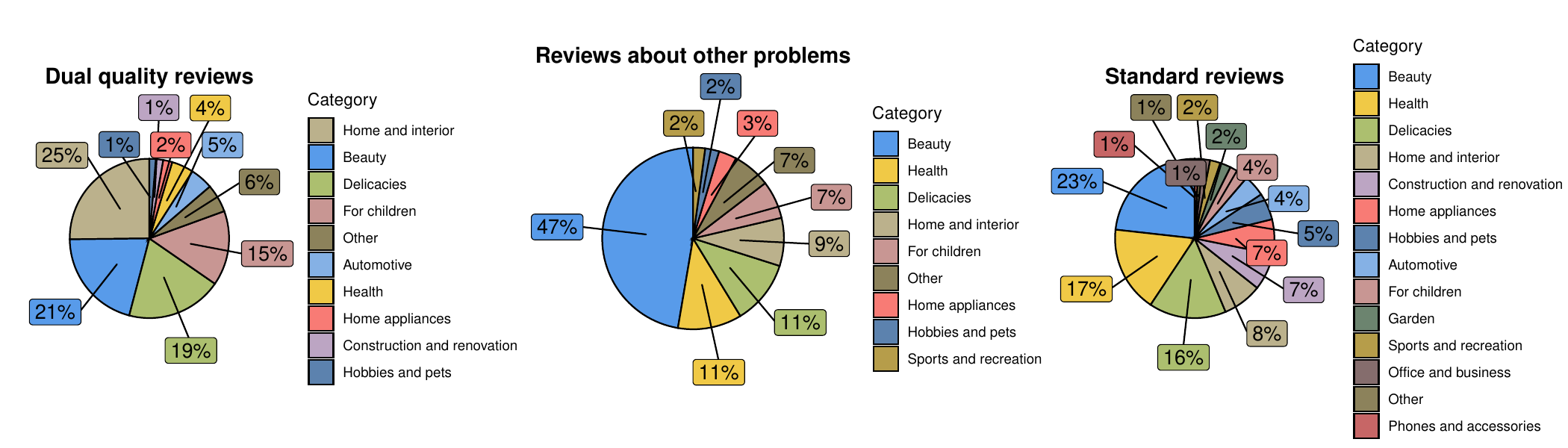}
\caption{Charts illustrating the distribution of product categories across various types of reviews.}
\label{fig:product_categories_break}
\end{figure*}

\section{DQ Dataset}

\subsection{Dataset Creation Methodology}
\label{subsec:dataset_creation}
In the first stage of our work, we collected a large dataset of reviews in Polish, sourced from the e-commerce platform CENEO\footnote{\url{https://www.ceneo.pl/}} and the discussion forum on beauty, makeup, and cosmetics, WIZAZ\footnote{\url{https://wizaz.pl/forum/}}. Our preliminary tests have shown that the problem of dual quality does not occur often in reviews, and thus randomly selecting a set of opinions and giving them to annotators is an inefficient approach to building a dataset. Therefore, we prepared a methodology to optimize this process, which consists of the following steps:

\noindent
{\bf \large \textcircled{\small 1}} Find dual quality reviews on the Internet by searching for publicly available articles that describe the problem of dual quality. Such articles often included examples of products along with the differences observed depending on the sales market, which we extracted. In addition, some articles had comment sections where people shared their experiences with the dual quality issue, which we also collected. In this way, we obtained \textbf{117} dual quality reviews.

\noindent
{\bf \large \textcircled{\small 2}} Randomly select \textbf{300} reviews from the CENEO / WIZAZ dataset as standard opinions that do not indicate a dual quality problem. These reviews have been verified to ensure that they are standard. Along with the examples obtained in step {\bf \large \textcircled{\small 1}}, these formed the base dataset.

\noindent
{\bf \large \textcircled{\small 3}} Train a model using a few-shot learning method to detect dual quality reviews based on the prepared base or an extended dataset (subsequent iterations). We adopted this approach due to the limited amount of training data. The model was implemented using the SetFit (Sentence Transformer Fine-tuning) framework \cite{setfit} and a sentence transformer for the Polish language st-polish-paraphrase-from-distilroberta\footnote{At the time of the dataset creation (beginning of 2023) it was the top Polish sentence transformer, as confirmed by \citet{arbuz}.}.

\noindent
{\bf \large \textcircled{\small 4}} Apply the model trained in step {\bf \large \textcircled{\small 3}} to all reviews of the CENEO / WIZAZ dataset. The results of the classification were sorted according to the probability returned by the model.

\noindent
{\bf \large \textcircled{\small 5}} Select up to \textbf{200}\footnote{Initially, many reviews were classified as dual quality, making a probability threshold unsuitable. Selecting 200 enabled swift human verification, speeding up subsequent iterations.} reviews with the highest probability of indicating a dual quality problem, which did not appear previously in the dataset. Then perform manual verification of the selected reviews. If a review did not indicate a dual quality issue, it was labeled as a standard review. During this step, we noticed that some reviews mentioned other problems, including, for example, the product being possibly counterfeit, deterioration in product quality over time, or the received product does not match the order. Annotators labeled such opinions as other problems and added additional information regarding the type of problem mentioned in the review. For training the model in step {\bf \large \textcircled{\small 3}}, the reviews labeled as other problems and standard were combined. The outcome of this step and the base dataset constituted the extended dataset.

\noindent
{\bf \large \textcircled{\small 6}} Return to step {\bf \large \textcircled{\small 3}} to increase the size of the dataset.

Steps {\bf \large \textcircled{\small 3}}, {\bf \large \textcircled{\small 4}}, and {\bf \large \textcircled{\small 5}} were repeated \textbf{7} times, allowing us to expand the base dataset with \textbf{1,303} examples (in last iteration only \textbf{103} new reviews were selected). We then applied the model, trained on the entire dataset prepared so far, to classify the reviews imported into the demo version of our system. Reviews were sourced from Polish and international e-commerce sites. Of these reviews, \textbf{237} were labeled as dual quality, which we manually verified and changed if necessary. As a result of the entire process described above, we obtained a DQ (\textbf{D}ual \textbf{Q}uality) dataset consisting of \textbf{1,957} unique examples. o ensure annotation accuracy, we conducted cross-validation and identified examples where the models were most often wrong. After verifying these errors, in \textbf{67 (3.4\%)} cases the label was incorrect and was changed. The whole above process is shown in Figure \ref{fig:dataset-creation}.

\subsection{Dataset Statistics}

The statistics of the DQ dataset are presented in Table \ref{tab:dq_dataset_stats}. The dataset consists of \textbf{1,957} records, of which \textbf{540} are labeled as dual quality, \textbf{281} as other problems, and the rest are standard opinions. Of the dual quality reviews, \textbf{107}\footnote{In the results of the final dataset verification, of the 117 dual quality reviews initially found, 10 were classified as standard.} were from the Internet, \textbf{265} from the CENEO / WIZAZ collection, and \textbf{168} from our demo system. The dataset is unbalanced, with over half of the reviews belong to the standard class. This characteristic was intentionally maintained because, in the real world, reviews on dual quality and other problems occur less frequently than others. 
For experimental purposes, the dataset was divided into three subsets: train, test and valid, containing \textbf{1,200 ($\sim$61\%)}, \textbf{500 ($\sim$26\%)}, and \textbf{257 ($\sim$13\%)} reviews, respectively. The review texts in the dataset consist of \textbf{261} characters and \textbf{41} words on average.

\begin{table}[h!]
\small
\begin{center}
{\renewcommand{\arraystretch}{1.1}%
\begin{tabular}{lr|rrr}
\hline
& \multicolumn{4}{c}{\bf \# reviews} \\
\cline{2-5}
 \bf label          & \bf all & \bf train & \bf test & \bf valid \\
\hline
 \bf dual quality   &  540  &     331 &    138 &      71  \\
 \bf other problems &   281 &     172 &     72 &      37  \\
 \bf standard       &  1136 &     697 &    290 &     149  \\
 \hline
 \bf total          &  1957 &    1200 &    500 &     257  \\
\hline
\end{tabular}}
\end{center}
\caption{DQ dataset statistics.}
\label{tab:dq_dataset_stats}
\end{table}

In addition, in Figure~\ref{fig:product_categories_break} we present pie charts depicting the distribution of product categories across various types of reviews\footnote{All product reviews categorized by product type reader may see in Figure~\ref{fig:product_categories}.}. A few interesting patterns in these distributions are worth describing. For instance, although \textit{Beauty}, \textit{Delicacies}, \textit{Health}, and \textit{Home \& Interior} are large categories overall, \textit{Home \& Interior} has an exceptionally high share among dual quality reviews (25\%, compared to 13\% overall), suggesting that this type of issue might be more commonly perceived in products related to household items. Similarly, \textit{For children} makes up only 7\% of all reviews but appears more prominently (15\%) in dual quality reviews. Meanwhile, \textit{Beauty} reviews account for nearly half (47\%) of the `other problems' category, indicating that consumers in that segment may encounter a broader range of product issues beyond dual quality concerns. 

\begin{table*}[h!]
\scriptsize
\centering
\begin{tabular}{llll|llll}
\hline
    & \multicolumn{3}{c}{\bf Dual Quality class} & \multicolumn{4}{c}{\bf All classes} \\
 \bf Method  & \bf Precision        & \bf Recall           & \bf F1         & \bf Accuracy        & \bf mPrecision        & \bf mRecall           & \bf mF1         \\
\hline
 Baseline                               & $42.4_{\pm0.0}$     & $84.8_{\pm0.0}$      & $56.5_{\pm0.0}$     & $55.2_{\pm0.0}$     & $37.8_{\pm0.0}$     & $46.5_{\pm0.0}$     & $39.5_{\pm0.0}$     \\
 \hline
 \multicolumn{8}{l}{\textbf{SetFit + sentence transformers}} \\
 \hline
 LaBSE                                       & $74.4_{\pm1.0}$                   & $71.4_{\pm2.2}$                    & $72.9_{\pm1.1}$                   & $77.7_{\pm0.5}$                   & $\bf 75.6_{\pm0.8}$               & $65.9_{\pm0.9}$                   & $68.4_{\pm0.7}$                   \\
 para-multi-mpnet-base-v2                    & $72.8_{\pm1.7}$                   & $66.4_{\pm2.4}$                    & $69.4_{\pm2.0}$                   & $75.9_{\pm1.4}$                   & $72.4_{\pm2.2}$                   & $66.8_{\pm2.5}$                   & $68.8_{\pm2.6}$                   \\
 para-multi-MiniLM-L12-v2                    & $69.4_{\pm2.2}$                   & $58.7_{\pm3.3}$                    & $63.6_{\pm2.7}$                   & $71.2_{\pm1.2}$                   & $65.8_{\pm1.3}$                   & $58.2_{\pm1.7}$                   & $60.2_{\pm1.7}$                   \\
 multi-e5-small                              & $68.7_{\pm1.6}$                   & $68.0_{\pm1.3}$                    & $68.3_{\pm0.8}$                   & $72.8_{\pm0.7}$                   & $70.4_{\pm0.8}$                   & $58.9_{\pm0.9}$                   & $60.3_{\pm1.3}$                   \\
 multi-e5-base                               & $72.2_{\pm1.2}$                   & $\bf 79.0_{\pm2.5}$                & $75.4_{\pm0.8}$                   & $77.4_{\pm1.0}$                   & $73.7_{\pm2.1}$                   & $67.6_{\pm1.8}$                   & $69.0_{\pm1.9}$                   \\
 multi-e5-large                              & $77.5_{\pm1.8}$                   & $76.8_{\pm3.4}$                    & $\bf 77.1_{\pm2.4}$               & $\bf 79.6_{\pm1.8}$               & $75.2_{\pm2.8}$                   & $71.2_{\pm2.2}$                   & $\bf 72.7_{\pm2.2}$               \\
 gte-multi-base                              & $73.4_{\pm1.1}$                   & $\bf 79.0_{\pm3.4}$                & $76.1_{\pm2.2}$                   & $78.6_{\pm0.8}$                   & $74.3_{\pm1.1}$                   & $69.4_{\pm2.0}$                   & $70.8_{\pm1.7}$                   \\
 st-polish-para-mpnet                        & $72.5_{\pm2.0}$                   & $71.7_{\pm3.3}$                    & $72.1_{\pm2.6}$                   & $76.6_{\pm1.1}$                   & $72.2_{\pm1.3}$                   & $68.1_{\pm2.1}$                   & $69.6_{\pm1.8}$                   \\
 st-polish-para-distilroberta                & $72.7_{\pm2.7}$                   & $69.1_{\pm2.7}$                    & $70.9_{\pm2.6}$                   & $75.7_{\pm0.7}$                   & $70.5_{\pm0.3}$                   & $68.1_{\pm1.6}$                   & $69.1_{\pm1.1}$                   \\
 mmlw-roberta-base                           & $\bf 77.9_{\pm0.8}$               & $73.6_{\pm1.6}$                    & $75.7_{\pm0.5}$                   & $78.6_{\pm0.6}$                   & $73.4_{\pm1.1}$                   & $71.9_{\pm1.0}$                   & $72.6_{\pm1.0}$                   \\
 mmlw-roberta-large                          & $76.0_{\pm1.9}$                   & $75.9_{\pm2.4}$                    & $75.9_{\pm2.0}$                   & $78.7_{\pm1.4}$                   & $72.7_{\pm1.8}$                   & $\bf 72.1_{\pm1.7}$               & $72.4_{\pm1.7}$                   \\
 \hline
 \multicolumn{8}{l}{\textbf{Transformer-based encoders}} \\
 \hline
 mBERT                                       & $64.8_{\pm2.7}$                   & $67.5_{\pm2.0}$                    & $66.1_{\pm1.6}$                   & $71.1_{\pm1.9}$                   & $62.5_{\pm9.4}$                   & $58.3_{\pm3.5}$                   & $58.6_{\pm5.5}$                   \\
 xlm-roberta-base                            & $60.7_{\pm1.5}$                   & $82.2_{\pm3.6}$                    & $69.8_{\pm1.1}$                   & $73.1_{\pm0.8}$                   & $70.6_{\pm1.1}$                   & $63.0_{\pm2.3}$                   & $62.8_{\pm2.5}$                   \\
 xlm-roberta-large                           & $78.3_{\pm3.0}$                   & $86.1_{\pm2.0}$                    & $\bf 82.0_{\pm1.5}$               & $82.0_{\pm1.2}$                   & $75.8_{\pm1.7}$                   & $\bf 76.4_{\pm1.6}$               & $75.9_{\pm1.6}$                   \\
 herbert-base-cased                          & $64.0_{\pm3.9}$                   & $77.8_{\pm3.3}$                    & $70.1_{\pm1.6}$                   & $73.3_{\pm0.2}$                   & $77.3_{\pm3.3}$                   & $59.9_{\pm2.3}$                   & $59.4_{\pm3.4}$                   \\
 herbert-large-cased                         & $81.5_{\pm2.5}$                   & $80.7_{\pm2.0}$                    & $81.1_{\pm1.5}$                   & \color{blue}{$\bf 82.4_{\pm1.1}$} & $77.6_{\pm1.4}$                   & $76.2_{\pm2.7}$                   & \color{blue}{$\bf 76.7_{\pm2.1}$} \\
 polish-roberta-base-v2                      & $66.4_{\pm3.0}$                   & $\bf 86.5_{\pm3.9}$                & $75.1_{\pm2.1}$                   & $75.4_{\pm1.5}$                   & $69.7_{\pm2.3}$                   & $67.2_{\pm1.9}$                   & $66.9_{\pm2.0}$                   \\
 polish-roberta-large-v2                     & $\bf 84.6_{\pm3.6}$               & $77.5_{\pm6.0}$                    & $80.7_{\pm2.9}$                   & $81.7_{\pm1.2}$                   & \color{blue}{$\bf 78.5_{\pm0.7}$} & $74.3_{\pm3.7}$                   & $75.8_{\pm2.5}$                   \\
 \hline
 \multicolumn{8}{l}{\textbf{LLMs}} \\
 \hline
  deepseek-v3 \textsubscript{zero-shot}       & $48.1_{\pm0.3}$                   & $90.6_{\pm1.2}$                    & $62.9_{\pm0.6}$                   & $49.5_{\pm0.4}$                   & $49.6_{\pm0.2}$                   & $47.9_{\pm0.4}$                   & $42.7_{\pm0.5}$                   \\
 deepseek-v3 \textsubscript{few-shot}        & $61.9_{\pm0.3}$                   & $96.1_{\pm0.3}$                    & $75.3_{\pm0.1}$                   & $59.0_{\pm0.2}$                   & $61.1_{\pm0.4}$                   & $63.7_{\pm0.4}$                   & $55.9_{\pm0.3}$                   \\
deepseek-v3 \textsubscript{zero-shot+inst.} & $84.7_{\pm1.3}$                   & $80.6_{\pm0.7}$                    & \color{blue}{$\bf 82.6_{\pm0.6}$} & $70.7_{\pm0.4}$                   & $70.4_{\pm0.6}$                   & $74.8_{\pm0.5}$                   & $68.7_{\pm0.4}$                   \\
 deepseek-v3 \textsubscript{few-shot+inst.}  & $79.7_{\pm0.9}$                   & $82.0_{\pm0.8}$                    & $80.9_{\pm0.9}$                   & $68.4_{\pm0.8}$                   & $70.1_{\pm0.6}$                   & $76.4_{\pm0.8}$                   & $67.4_{\pm0.8}$                   \\
 gpt-4o \textsubscript{zero-shot}            & $42.8_{\pm0.2}$                   & \color{blue}{$\bf 100.0_{\pm0.0}$} & $60.0_{\pm0.2}$                   & $47.6_{\pm0.3}$                   & $49.8_{\pm0.2}$                   & $46.8_{\pm0.3}$                   & $38.8_{\pm0.3}$                   \\
 gpt-4o \textsubscript{few-shot}             & $60.3_{\pm0.2}$                   & $98.8_{\pm0.3}$                    & $74.9_{\pm0.3}$                   & $57.5_{\pm0.2}$                   & $62.1_{\pm0.1}$                   & $66.5_{\pm0.3}$                   & $55.5_{\pm0.3}$                   \\
 gpt-4o \textsubscript{zero-shot+inst.}      & $85.7_{\pm0.4}$                   & $76.7_{\pm0.8}$                    & $80.9_{\pm0.6}$                   & $\bf 75.0_{\pm0.2}$               & $\bf 73.4_{\pm0.2}$               & \color{blue}{$\bf 79.0_{\pm0.3}$} & $\bf 72.5_{\pm0.2}$               \\
 gpt-4o \textsubscript{few-shot+inst.}       & \color{blue}{$\bf 86.0_{\pm1.9}$} & $75.1_{\pm0.7}$                    & $80.1_{\pm0.6}$                   & $68.5_{\pm0.3}$                   & $72.3_{\pm0.5}$                   & $76.5_{\pm0.2}$                   & $67.7_{\pm0.3}$                   \\
\hline
\end{tabular}
\caption{\label{tab:main-results}
Average scores with standard deviation for all evaluated methods. The Precision, Recall, and F1 metrics were calculated considering only the dual quality class; the other metrics were for all classes, with 'm' as the macro average. Bold values indicate the highest scores for the type of method, and blue highlights the highest scores for each metric.} 
\end{table*}

\section{Experiments}

\subsection{Experimental Setup}
\label{sec:experimental_setup}
The problem was defined as a three-class classification (see Table~\ref{tab:dq_dataset_stats}). Evaluation of various methods was performed on a test set. The training set and the validation set were used for approaches that required training/fine-tuning. Each experiment was repeated five times\footnote{This rule was not applied to Baseline, which is deterministic, and successive runs always produce the same result.}, setting a different seed value (if applicable), and the results presented in the tables are average values. 

\subsection{Methods}

\textbf{Baseline} is a naive method of assigning a dual quality class to a review if there are references to another country in the text.

\noindent
\textbf{SetFit + sentence transformers} is an approach in which a sentence transformer model is first fine-tuned using contrastive learning and then used as text embedding for a logistic regression model. In the experiments, we used sentence transformers previously tested on the PL-MTEB benchmark by \citet{pl-mteb}. We selected seven multilingual models namely: LaBSE~\cite{feng2022}, paraphrase-multilingual-mpnet-base-v2, paraphrase-multilingual-MiniLM-L12-v2 \cite{sentence-bert},  three e5 models~\cite{wang2024multilingual} and mGTE~\cite{mgte}. Additionally, we choose four sentence-transformer models dedicated to the Polish language: st-polish-paraphrase-from-mpnet, st-polish-paraphrase-from-distilroberta \cite{arbuz} and two mmlw models \cite{pirb}.

\noindent
\textbf{Transformer-based encoders}  involves training pre-trained language model with classification head on top (a linear layer on top of the pooled output). We included evaluations of multilingual BERT (mBERT)~\cite{devlin2019}, multilingual XLM-RoBERTa~\cite{conneau2020}, and models specifically trained for Polish, such as HerBERT~\cite{mroczkowski2021} and Polish RoBERTa~\cite{dadas2020pretraining}. 

\noindent
\textbf{LLMs}
Advanced frontier models such as DeepSeek~\cite{deepseekai2025,deepseekai2024} and GPT-4o~\cite{openai2024} were selected to evaluate how effectively cutting-edge LLMs handle dual quality review detection tasks under different prompting scenarios, including zero-shot and few-shot configurations, both with and without additional instruction (see more details about used prompts in Table~\ref{tab:promptspl}).

\begin{figure*}[!htb]
\centering
\includegraphics[scale=0.43]{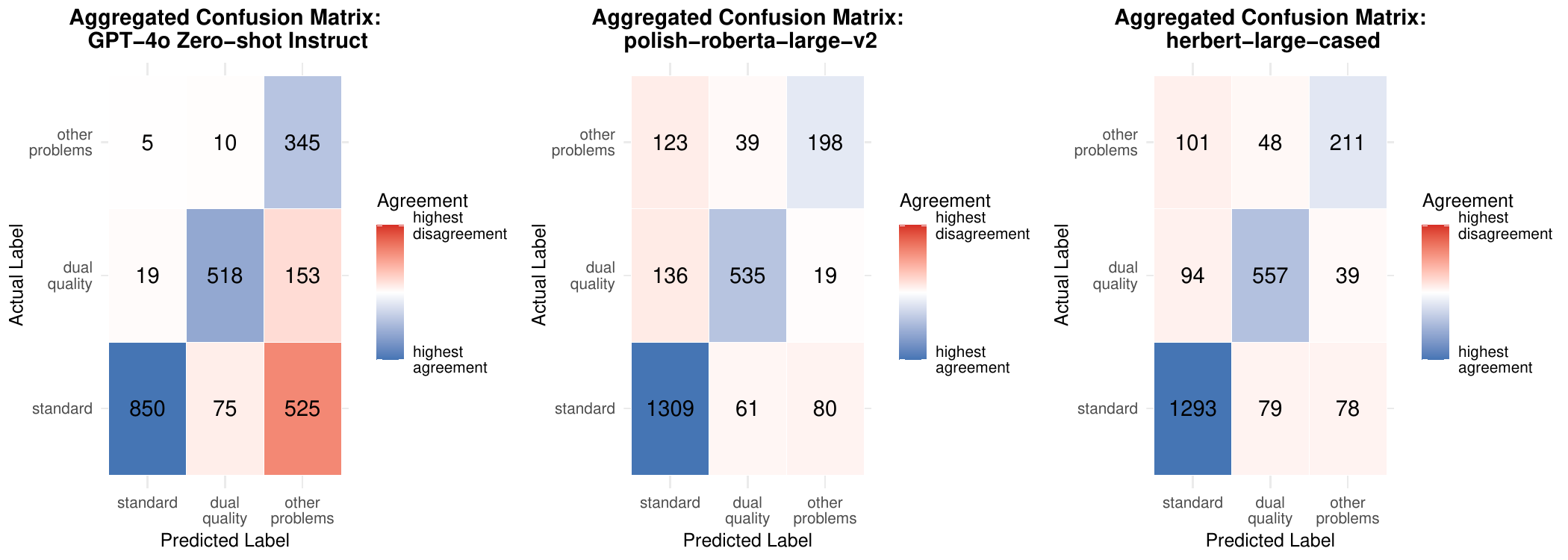}
\caption{Confusion matrices aggregated from five experiments for selected models.}
\label{fig:hm1}
\end{figure*}

\subsection{Main Results}\label{sec:mainres}

The experimental results from Table~\ref{tab:main-results} clearly indicate notable differences among the three groups of tested models. Sentence-transformer models using SetFit generally achieved moderate precision scores (around 70-77\%), suggesting that compressing sentence semantics into a single vector might result in information loss or inadequate semantic representation. Transformer-based encoders, particularly the larger, language-specific models such as polish-roberta-large-v2 (84.6\%) and herbert-large-cased (81.5\%), exhibited significantly stronger performance, comparable even with state-of-the-art conversational large language models (LLMs). Among LLMs, instructive prompting strategies (providing clear definitions of classes without explicit examples) improved performance, with the best precision results of 86\% and 85.7\% achieved by GPT-4o models with and without examples, respectively. It should be noted that the GPT-4o model with zero-shot instr. prompt achieved very good results for other measures as well.  Interestingly, explicit few-shot examples sometimes distort the models and reduce detection efficiency overall. This may suggest that the chosen examples may not be representative and therefore helpful.  


\subsection{Errors Analysis}\label{sec:erranalys}

We conducted a detailed error analysis for selected models using classification confusion matrices visualized through heat maps. Specifically, we selected three representative models: GPT-4o (zero-shot+inst.), polish-roberta-large-v2 and herbert-large-cased.
Figure~\ref{fig:hm1} shows that the GPT-4o model exhibits substantial confusion between standard and `other problems' reviews, while errors between standard and dual quality are less frequent. The polish-roberta-large-v2 model frequently identifies the standard reviews, achieving high accuracy for this category, but often misclassifies dual quality opinions as standard. 
Model herbert-large-cased often recognizes the dual quality reviews, achieving a high detection rate but also producing the most false positives for this class.
Additional comparative analyses are presented in Figure~\ref{fig:hm2} and Figure~\ref{fig:hm3}. 

\subsection{Robustness}\label{sec:secrobustness}

As an additional experiment, we verified robustness of selected models, i.e., whether a slight change in the text, which does not significantly affect its meaning, can change the model's decision. We generated five additional test sets, which resulted from modifications to the original test set. The modifications are described in Table \ref{tab:robust_desc}.
We tested three selected models, the results are shown in Table \ref{tab:robust}. The percentage of differences in predictions was between 2.6 and 5.0. More often, larger text modifications like pl\_chars influenced the change in decision.

\begin{table}[h!]
\scriptsize
\centering
\begin{tabular}{p{0.2\columnwidth}p{0.7\columnwidth}}
\hline
 \bf Name         &  \bf Description  \\
\hline
 period           & Remove (if present) or add (if absent) a period at the end of the review. \\ 
 first\_letter    & Change the capitalization of the first letter of the first word in the review. If the first word is written in uppercase, change it to lowercase. \\
 lower            & Change text of the review to lowercase. \\
 pl\_chars   & Replace the Polish characters \textit{ą, ę, ć, ł, ń, ó, ż, ź} with their corresponding Latin alphabet characters, i.e., \textit{a, e, c, l, n, o, z}. \\
 pl\_chars\_once   & The operation is the same as pl\_chars, except that each letter can be changed once. \\
\hline
\end{tabular}
\caption{Descriptions of modifications applied to the test set for robustness verification.}
\label{tab:robust_desc}
\end{table}

\begin{table}[h!]
\scriptsize
\centering
\begin{tabular}{llll}
\hline
 \bf Modification & \bf \rot{gpt-4o}   & \bf \rot{polish-roberta} & \bf \rot{herbert}  \\
\hline
period          & $4.0_{\pm0.0}$                            & $4.2_{\pm1.0}$            & $5.0_{\pm0.9}$        \\
 first\_letter   & $4.0_{\pm0.0}$                            & $2.8_{\pm0.7}$            & $2.6_{\pm0.8}$        \\
 lower           & $5.0_{\pm0.0}$                            & $4.6_{\pm0.5}$            & $4.2_{\pm0.7}$        \\
 pl\_chars       & $5.0_{\pm0.0}$                            & $4.6_{\pm1.2}$            & $4.6_{\pm0.8}$        \\
 pl\_chars\_once & $4.0_{\pm0.0}$                            & $4.0_{\pm1.4}$            & $3.6_{\pm0.8}$        \\ 
\hline
\end{tabular}
\caption{Robustness verification results for GPT-4o (zero-shot+inst.), polish-roberta-large-v2 and herbert-large-cased. The values are the average and standard deviation of the model's decision disagreement for the original and modified reviews. To ensure consistent behavior in the GPT-4o model, we set the temperature to 0.0, resulting in a standard deviation of 0.0 across runs.}
\label{tab:robust}
\end{table}

\begin{table*}[h!]
\scriptsize
\centering
\begin{tabular}{llll|llll}
\hline
& \multicolumn{3}{c}{\bf Dual Quality class} & \multicolumn{4}{c}{\bf All classes} \\
 \bf Method  & \bf Precision        & \bf Recall           & \bf F1         & \bf Accuracy        & \bf mPrecision        & \bf mRecall           & \bf mF1         \\
\hline
 \multicolumn{8}{l}{\textbf{Transformer-based encoders}} \\
 \hline
 xlm-roberta-base                            & $69.5_{\pm2.3}$                   & $\bf 66.9_{\pm6.8}$               & $67.9_{\pm2.9}$                   & $\bf 73.0_{\pm1.0}$               & $55.5_{\pm1.1}$                   & $55.1_{\pm2.1}$                   & $55.0_{\pm1.7}$                   \\
 xlm-roberta-large                           & $\bf 84.8_{\pm3.8}$               & $63.1_{\pm4.8}$                   & \color{blue}{$\bf 72.3_{\pm4.0}$} & $72.6_{\pm2.7}$                   & $\bf 60.1_{\pm2.7}$               & $\bf 56.7_{\pm3.9}$               & \color{blue}{$\bf 57.5_{\pm3.3}$} \\
  \hline
 \multicolumn{8}{l}{\textbf{LLMs}} \\
 \hline
 deepseek-v3 \textsubscript{zero-shot+inst.} & $85.9_{\pm1.8}$                   & $52.3_{\pm0.8}$                   & $65.0_{\pm0.3}$                   & $49.5_{\pm0.7}$                   & $63.4_{\pm1.3}$                   & \color{blue}{$\bf 58.7_{\pm1.0}$} & $49.1_{\pm0.7}$                   \\
 deepseek-v3 \textsubscript{few-shot+inst.}  & \color{blue}{$\bf 91.9_{\pm4.8}$} & $50.6_{\pm0.8}$                   & $65.2_{\pm1.8}$                   & $44.3_{\pm0.9}$                   & \color{blue}{$\bf 65.6_{\pm2.2}$} & $56.2_{\pm1.2}$                   & $46.1_{\pm1.0}$                   \\
 gpt-4o \textsubscript{zero-shot+inst.}      & $85.3_{\pm1.3}$                   & $46.6_{\pm0.0}$                   & $60.2_{\pm0.3}$                   & $\bf 52.6_{\pm0.6}$               & $62.3_{\pm0.3}$                   & $57.1_{\pm0.3}$                   & $\bf 49.6_{\pm0.3}$               \\
 gpt-4o \textsubscript{few-shot+inst.}       & $80.2_{\pm1.1}$                   & $46.6_{\pm0.0}$                   & $58.9_{\pm0.3}$                   & $41.6_{\pm0.6}$                   & $61.4_{\pm0.5}$                   & $50.2_{\pm1.0}$                   & $42.7_{\pm0.5}$                   \\
\hline
\end{tabular}
\caption{\label{tab:multi-results}
Evaluation results for selected models on a multilingual dataset.}
\end{table*}

\subsection{Multilingual Transfer}
To verify generalizability across markets and languages, we also explored multilingual transfer capabilities of our solution. For this purpose, we created a multilingual subset of reviews in English, German, and French (200,000 reviews for each language) selected from the AMAZON \cite{keung-etal-2020-multilingual} dataset and our demo system. Next, we trained SetFit with paraphrase-multilingual-mpnet-base-v2\footnote{One of the top multilingual sentence transformer at that time (2023).} on the DQ dataset, and applied it to these reviews. Then we selected 500 AMAZON reviews and 200 reviews from demo system with the highest dual quality scores. Manual verification showed that most were actually standard, so we randomly limited standard reviews to 130, yielding \textbf{206} final examples (\textbf{58} dual quality, \textbf{18} other problems, \textbf{130} standard). The dataset thus prepared was used as a multilingual test set. We conducted an experiment in which we tested methods based on multilingual models trained as in Section \ref{sec:experimental_setup} on the Polish training subset or, in the case of LLMs, using the same prompts. The results for the selected models are presented in Table \ref{tab:multi-results}. Considering the precision of the classifier, the highest score was achieved by the DeepSeek-V3 (91.9\%) model, interestingly in this case, adding examples to the instructions in the prompt gave a higher score. Of the group of transformer-based encoders, the highest score was achieved by xlm-roberta-large (84.8\%). Although the difference in performance on the basis of precision is significant, it is important to note the low values of the recall measure for LLMs, compared to encoders. All results for this experiment are available in Table~ \ref{tab:multi-results-full}.


\section{Deployment and Practical Considerations}
During the evaluation, a key objective was to achieve high precision, thereby minimizing the number of false positive recommendations. Since each flagged instance undergoes final verification by a human analyst, the primary goal is to reduce the analyst's workload by minimizing the number of irrelevant alerts. This approach accepts the possibility of missing some true dual quality cases (i.e., allowing for a certain level of false negatives) in favor of ensuring that the identified cases are highly likely to be accurate. A product with several dual quality reviews will be selected for further analysis to verify whether this issue genuinely exists in its case.

The proposed solution is implemented as a standalone service within a local infrastructure and is exclusively dedicated to UOKiK employees (Poland's Office of Competition and Consumer Protection). The system is currently not accessible to the public or external users. Although the system can analyze multilingual content, the current deployment prioritizes support for the Polish language to align with the context of Polish consumers and UOKiK's mandate within the Polish market.

Given the results of the evaluation and the above assumptions, we would recommend using the polish-robert-large-v2 model for a production deployment. Selecting the locally deployable model presents a pragmatic and efficient choice, particularly when minimizing external dependencies and ensuring consistent, low-latency inference.
It should be noted that this language-specific component is modular; for deployment within other European consumer protection agencies analogous to UOKiK, the model could be readily substituted with an equivalent model fine-tuned for the respective national language (e.g., a German BERT for a German institution) or multilingual model like XLM-RoBERTa.


\section{Conclusion}

In this work, we presented the entire process of preparing a solution for detecting the problem of dual quality based on product reviews. Our three key findings are: First, mentions of dual quality in product reviews are rare, in our case appearing only a few hundred times. Second, smaller language-specific transformer-based encoders finetuned for the task perform comparably to larger LLMs. Finally, including examples in prompts for LLMs can degrade performance compared to using only task-specific instructions.

\section*{Acknowledgments}
Project co-financed/financed by the National Centre for Research and Development (\url{https://www.gov.pl/web/ncbr-en}) under the programme Infostrateg III.

\bibliography{acl_latext}

\begin{thebibliography}{45}
\providecommand{\natexlab}[1]{#1}

\bibitem[{Avigdor et~al.(2023)Avigdor, Horowitz, Raviv, and Yanovsky~Daye}]{avigdor2023}
Noa Avigdor, Guy Horowitz, Ariel Raviv, and Stav Yanovsky~Daye. 2023.
\newblock \href {https://doi.org/10.18653/v1/2023.acl-industry.30} {Consistent text categorization using data augmentation in e-commerce}.
\newblock In \emph{Proceedings of the 61st Annual Meeting of the Association for Computational Linguistics (Volume 5: Industry Track)}, pages 313--321, Toronto, Canada. Association for Computational Linguistics.

\bibitem[{Bartkova and Sirotiaková(2021)}]{2021blucia}
Lucia Bartkova and Mária Sirotiaková. 2021.
\newblock \href {https://doi.org/10.1051/shsconf/20219206001} {Dual quality and its influence on consumer behaviour according to the income}.
\newblock \emph{SHS Web of Conferences}, 92.

\bibitem[{Bartkova and Veselovska(2023)}]{2023lenka}
Lucia Bartkova and Lenka Veselovska. 2023.
\newblock \href {https://doi.org/10.21272/mmi.2023.1-16} {Does dual quality of products in the european union truly bother consumers?}
\newblock \emph{Marketing and Management of Innovations}, 14.

\bibitem[{Bartkova et~al.(2021)Bartkova, Veselovska, Sramkova, and Zavadsky}]{2021lucia}
Lucia Bartkova, Lenka Veselovska, Marianna Sramkova, and Jan Zavadsky. 2021.
\newblock \href {https://doi.org/10.21272/mmi.2021.1-18} {Dual quality of products: myths and facts through the opinions of millennial consumers}.
\newblock \emph{Marketing and Management of Innovations}.

\bibitem[{Bartková and Veselovská(2024)}]{2024lucia}
L.~Bartková and L.~Veselovská. 2024.
\newblock \href {https://doi.org/10.1016/j.iimb.2024.05.001} {Consumer behaviour under dual quality of products: Does testing reveal what consumers experience?}
\newblock \emph{IIMB Management Review}, 36:171--184.

\bibitem[{Bartková(2019)}]{2019lucia}
Lucia Bartková. 2019.
\newblock \href {https://doi.org/10.21511/ppm.17(3).2019.31} {How do consumers perceive the dual quality of goods and its economic aspects in the european union? an empirical study}.
\newblock \emph{Problems and Perspectives in Management}, 17.

\bibitem[{Bartková et~al.(2018)Bartková, Veselovská, and Zimermanová}]{2018lucia}
Lucia Bartková, Lenka Veselovská, and Katarína Zimermanová. 2018.
\newblock Possible solutions to dual quality of products in the european union.
\newblock \emph{Scientific Papers of the University of Pardubice, Series D: Faculty of Economics and Administration}, 26.

\bibitem[{Botunac et~al.(2024)Botunac, Bakarić, and Matetić}]{ivebotunax2024}
I.~Botunac, M.~Brkić Bakarić, and M.~Matetić. 2024.
\newblock \href {https://doi.org/10.3390/app14146254} {Comparing fine-tuning and prompt engineering for multi-class classification in hospitality review analysis}.
\newblock \emph{Applied Sciences (Switzerland)}, 14.

\bibitem[{{Chambers}()}]{2025Chambers}
{Chambers}.
\newblock {Dual Quality of Food Products}.
\newblock \url{https://chambers.com/legal-trends/dual-quality-of-food-products}.
\newblock [Online; accessed 06-March-2025].

\bibitem[{Commission(2018)}]{2018ECNewDeal}
European Commission. 2018.
\newblock {Dual quality of food: European Commission releases common testing methodology}.
\newblock \url{https://ec.europa.eu/commission/presscorner/detail/en/ip_18_4122}.
\newblock [Online; accessed 06-March-2025].

\bibitem[{Conneau et~al.(2020)Conneau, Khandelwal, Goyal, Chaudhary, Wenzek, Guzm{\'a}n, Grave, Ott, Zettlemoyer, and Stoyanov}]{conneau2020}
Alexis Conneau, Kartikay Khandelwal, Naman Goyal, Vishrav Chaudhary, Guillaume Wenzek, Francisco Guzm{\'a}n, Edouard Grave, Myle Ott, Luke Zettlemoyer, and Veselin Stoyanov. 2020.
\newblock \href {https://doi.org/10.18653/v1/2020.acl-main.747} {Unsupervised cross-lingual representation learning at scale}.
\newblock In \emph{Proceedings of the 58th Annual Meeting of the Association for Computational Linguistics}, pages 8440--8451, Online. Association for Computational Linguistics.

\bibitem[{Dadas et~al.(2020)Dadas, Pere{\l}kiewicz, and Po{\'{s}}wiata}]{dadas2020pretraining}
S{\l}awomir Dadas, Micha{\l} Pere{\l}kiewicz, and Rafa{\l} Po{\'{s}}wiata. 2020.
\newblock Pre-training polish transformer-based language models at scale.
\newblock In \emph{Artificial Intelligence and Soft Computing}, pages 301--314. Springer International Publishing.

\bibitem[{Dadas et~al.(2024{\natexlab{a}})Dadas, Pere{\l}kiewicz, and Po{\'s}wiata}]{pirb}
Slawomir Dadas, Micha{\l} Pere{\l}kiewicz, and Rafa{\l} Po{\'s}wiata. 2024{\natexlab{a}}.
\newblock \href {https://aclanthology.org/2024.lrec-main.1117/} {{PIRB}: A comprehensive benchmark of {P}olish dense and hybrid text retrieval methods}.
\newblock In \emph{Proceedings of the 2024 Joint International Conference on Computational Linguistics, Language Resources and Evaluation (LREC-COLING 2024)}, pages 12761--12774, Torino, Italia. ELRA and ICCL.

\bibitem[{Dadas et~al.(2024{\natexlab{b}})Dadas, Kozłowski, Poświata, Perełkiewicz, Białas, and Grębowiec}]{arbuz}
Sławomir Dadas, Marek Kozłowski, Rafał Poświata, Michał Perełkiewicz, Marcin Białas, and Małgorzata Grębowiec. 2024{\natexlab{b}}.
\newblock \href {https://doi.org/10.1007/s10506-024-09408-8} {A support system for the detection of abusive clauses in b2c contracts}.
\newblock \emph{Artificial Intelligence and Law}.

\bibitem[{DeepSeek-AI et~al.(2024)DeepSeek-AI, :, Bi, Chen, Chen, Chen, Dai, Deng, Ding, Dong, Du, Fu, Gao, Gao, Gao, Ge, Guan, Guo, Guo, Hao, Hao, He, Hu, Huang, Li, Li, Li, Li, Li, Liang, Lin, Liu, Liu, Liu, Liu, Liu, Liu, Lu, Lu, Luo, Ma, Nie, Pei, Piao, Qiu, Qu, Ren, Ren, Ruan, Sha, Shao, Song, Su, Sun, Sun, Tang, Wang, Wang, Wang, Wang, Wang, Wu, Wu, Xie, Xie, Xie, Xiong, Xu, Xu, Xu, Yang, You, Yu, Yu, Zhang, Zhang, Zhang, Zhang, Zhang, Zhang, Zhang, Zhang, Zhao, Zhao, Zhou, Zhou, Zhu, and Zou}]{deepseekai2024}
DeepSeek-AI, :, Xiao Bi, Deli Chen, Guanting Chen, Shanhuang Chen, Damai Dai, Chengqi Deng, Honghui Ding, Kai Dong, Qiushi Du, Zhe Fu, Huazuo Gao, Kaige Gao, Wenjun Gao, Ruiqi Ge, Kang Guan, Daya Guo, Jianzhong Guo, Guangbo Hao, Zhewen Hao, Ying He, Wenjie Hu, Panpan Huang, Erhang Li, Guowei Li, Jiashi Li, Yao Li, Y.~K. Li, Wenfeng Liang, Fangyun Lin, A.~X. Liu, Bo~Liu, Wen Liu, Xiaodong Liu, Xin Liu, Yiyuan Liu, Haoyu Lu, Shanghao Lu, Fuli Luo, Shirong Ma, Xiaotao Nie, Tian Pei, Yishi Piao, Junjie Qiu, Hui Qu, Tongzheng Ren, Zehui Ren, Chong Ruan, Zhangli Sha, Zhihong Shao, Junxiao Song, Xuecheng Su, Jingxiang Sun, Yaofeng Sun, Minghui Tang, Bingxuan Wang, Peiyi Wang, Shiyu Wang, Yaohui Wang, Yongji Wang, Tong Wu, Y.~Wu, Xin Xie, Zhenda Xie, Ziwei Xie, Yiliang Xiong, Hanwei Xu, R.~X. Xu, Yanhong Xu, Dejian Yang, Yuxiang You, Shuiping Yu, Xingkai Yu, B.~Zhang, Haowei Zhang, Lecong Zhang, Liyue Zhang, Mingchuan Zhang, Minghua Zhang, Wentao Zhang, Yichao Zhang, Chenggang Zhao, Yao Zhao, Shangyan Zhou, Shunfeng
  Zhou, Qihao Zhu, and Yuheng Zou. 2024.
\newblock \href {https://arxiv.org/abs/2401.02954} {Deepseek llm: Scaling open-source language models with longtermism}.
\newblock \emph{Preprint}, arXiv:2401.02954.

\bibitem[{DeepSeek-AI et~al.(2025)DeepSeek-AI, Liu, Feng, Xue, Wang, Wu, Lu, Zhao, Deng, Zhang, Ruan, Dai, Guo, Yang, Chen, Ji, Li, Lin, Dai, Luo, Hao, Chen, Li, Zhang, Bao, Xu, Wang, Zhang, Ding, Xin, Gao, Li, Qu, Cai, Liang, Guo, Ni, Li, Wang, Chen, Chen, Yuan, Qiu, Li, Song, Dong, Hu, Gao, Guan, Huang, Yu, Wang, Zhang, Xu, Xia, Zhao, Wang, Zhang, Li, Wang, Zhang, Zhang, Tang, Li, Tian, Huang, Wang, Zhang, Wang, Zhu, Chen, Du, Chen, Jin, Ge, Zhang, Pan, Wang, Xu, Zhang, Chen, Li, Lu, Zhou, Chen, Wu, Ye, Ye, Ma, Wang, Zhou, Yu, Zhou, Pan, Wang, Yun, Pei, Sun, Xiao, Zeng, Zhao, An, Liu, Liang, Gao, Yu, Zhang, Li, Jin, Wang, Bi, Liu, Wang, Shen, Chen, Zhang, Chen, Nie, Sun, Wang, Cheng, Liu, Xie, Liu, Yu, Song, Shan, Zhou, Yang, Li, Su, Lin, Li, Wang, Wei, Zhu, Zhang, Xu, Xu, Huang, Li, Zhao, Sun, Li, Wang, Yu, Zheng, Zhang, Shi, Xiong, He, Tang, Piao, Wang, Tan, Ma, Liu, Guo, Wu, Ou, Zhu, Wang, Gong, Zou, He, Zha, Xiong, Ma, Yan, Luo, You, Liu, Zhou, Wu, Ren, Ren, Sha, Fu, Xu, Huang, Zhang, Xie, Zhang, Hao,
  Gou, Ma, Yan, Shao, Xu, Wu, Zhang, Li, Gu, Zhu, Liu, Li, Xie, Song, Gao, and Pan}]{deepseekai2025}
DeepSeek-AI, Aixin Liu, Bei Feng, Bing Xue, Bingxuan Wang, Bochao Wu, Chengda Lu, Chenggang Zhao, Chengqi Deng, Chenyu Zhang, Chong Ruan, Damai Dai, Daya Guo, Dejian Yang, Deli Chen, Dongjie Ji, Erhang Li, Fangyun Lin, Fucong Dai, Fuli Luo, Guangbo Hao, Guanting Chen, Guowei Li, H.~Zhang, Han Bao, Hanwei Xu, Haocheng Wang, Haowei Zhang, Honghui Ding, Huajian Xin, Huazuo Gao, Hui Li, Hui Qu, J.~L. Cai, Jian Liang, Jianzhong Guo, Jiaqi Ni, Jiashi Li, Jiawei Wang, Jin Chen, Jingchang Chen, Jingyang Yuan, Junjie Qiu, Junlong Li, Junxiao Song, Kai Dong, Kai Hu, Kaige Gao, Kang Guan, Kexin Huang, Kuai Yu, Lean Wang, Lecong Zhang, Lei Xu, Leyi Xia, Liang Zhao, Litong Wang, Liyue Zhang, Meng Li, Miaojun Wang, Mingchuan Zhang, Minghua Zhang, Minghui Tang, Mingming Li, Ning Tian, Panpan Huang, Peiyi Wang, Peng Zhang, Qiancheng Wang, Qihao Zhu, Qinyu Chen, Qiushi Du, R.~J. Chen, R.~L. Jin, Ruiqi Ge, Ruisong Zhang, Ruizhe Pan, Runji Wang, Runxin Xu, Ruoyu Zhang, Ruyi Chen, S.~S. Li, Shanghao Lu, Shangyan Zhou, Shanhuang
  Chen, Shaoqing Wu, Shengfeng Ye, Shengfeng Ye, Shirong Ma, Shiyu Wang, Shuang Zhou, Shuiping Yu, Shunfeng Zhou, Shuting Pan, T.~Wang, Tao Yun, Tian Pei, Tianyu Sun, W.~L. Xiao, Wangding Zeng, Wanjia Zhao, Wei An, Wen Liu, Wenfeng Liang, Wenjun Gao, Wenqin Yu, Wentao Zhang, X.~Q. Li, Xiangyue Jin, Xianzu Wang, Xiao Bi, Xiaodong Liu, Xiaohan Wang, Xiaojin Shen, Xiaokang Chen, Xiaokang Zhang, Xiaosha Chen, Xiaotao Nie, Xiaowen Sun, Xiaoxiang Wang, Xin Cheng, Xin Liu, Xin Xie, Xingchao Liu, Xingkai Yu, Xinnan Song, Xinxia Shan, Xinyi Zhou, Xinyu Yang, Xinyuan Li, Xuecheng Su, Xuheng Lin, Y.~K. Li, Y.~Q. Wang, Y.~X. Wei, Y.~X. Zhu, Yang Zhang, Yanhong Xu, Yanhong Xu, Yanping Huang, Yao Li, Yao Zhao, Yaofeng Sun, Yaohui Li, Yaohui Wang, Yi~Yu, Yi~Zheng, Yichao Zhang, Yifan Shi, Yiliang Xiong, Ying He, Ying Tang, Yishi Piao, Yisong Wang, Yixuan Tan, Yiyang Ma, Yiyuan Liu, Yongqiang Guo, Yu~Wu, Yuan Ou, Yuchen Zhu, Yuduan Wang, Yue Gong, Yuheng Zou, Yujia He, Yukun Zha, Yunfan Xiong, Yunxian Ma, Yuting Yan, Yuxiang
  Luo, Yuxiang You, Yuxuan Liu, Yuyang Zhou, Z.~F. Wu, Z.~Z. Ren, Zehui Ren, Zhangli Sha, Zhe Fu, Zhean Xu, Zhen Huang, Zhen Zhang, Zhenda Xie, Zhengyan Zhang, Zhewen Hao, Zhibin Gou, Zhicheng Ma, Zhigang Yan, Zhihong Shao, Zhipeng Xu, Zhiyu Wu, Zhongyu Zhang, Zhuoshu Li, Zihui Gu, Zijia Zhu, Zijun Liu, Zilin Li, Ziwei Xie, Ziyang Song, Ziyi Gao, and Zizheng Pan. 2025.
\newblock \href {https://arxiv.org/abs/2412.19437} {Deepseek-v3 technical report}.
\newblock \emph{Preprint}, arXiv:2412.19437.

\bibitem[{Devlin et~al.(2019)Devlin, Chang, Lee, and Toutanova}]{devlin2019}
Jacob Devlin, Ming-Wei Chang, Kenton Lee, and Kristina Toutanova. 2019.
\newblock \href {https://doi.org/10.18653/v1/N19-1423} {{BERT}: Pre-training of deep bidirectional transformers for language understanding}.
\newblock In \emph{Proceedings of the 2019 Conference of the North {A}merican Chapter of the Association for Computational Linguistics: Human Language Technologies, Volume 1 (Long and Short Papers)}, pages 4171--4186, Minneapolis, Minnesota. Association for Computational Linguistics.

\bibitem[{{EU Monitor}()}]{2025EUMonitor}
{EU Monitor}.
\newblock {The better enforcement and modernisation of Union consumer protection rules}.
\newblock \url{https://www.eumonitor.eu/9353000/1/j4nvhdfcs8bljza_j9vvik7m1c3gyxp/vme85bbfssxo}.
\newblock [Online; accessed 06-March-2025].

\bibitem[{European~Commission(2023)}]{2023ECJRC}
Joint Research~Centre European~Commission. 2023.
\newblock {Same pack, different ingredients: Is dual quality down-branded in EU food?}
\newblock \url{https://joint-research-centre.ec.europa.eu/jrc-news-and-updates/same-pack-different-ingredients-dual-quality-down-branded-eu-food-2023-07-24_en}.
\newblock [Online; accessed 06-March-2025].

\bibitem[{European~Parliament(2017)}]{2017EPRS}
European Parliamentary Research Service~(EPRS) European~Parliament. 2017.
\newblock {European Commission guidelines on dual quality of branded food products}.
\newblock \url{https://www.europarl.europa.eu/RegData/etudes/BRIE/2017/608804/EPRS_BRI%282017%29608804_EN.pdf}.
\newblock [PDF; accessed 06-March-2025].

\bibitem[{European~Parliament(2019)}]{2019EPRS}
European Parliamentary Research Service~(EPRS) European~Parliament. 2019.
\newblock {Dual quality of products – State of play}.
\newblock \url{https://www.europarl.europa.eu/RegData/etudes/BRIE/2019/644192/EPRS_BRI(2019)644192_EN.pdf}.
\newblock [Online; accessed 06-March-2025].

\bibitem[{Feng et~al.(2022)Feng, Yang, Cer, Arivazhagan, and Wang}]{feng2022}
Fangxiaoyu Feng, Yinfei Yang, Daniel Cer, Naveen Arivazhagan, and Wei Wang. 2022.
\newblock \href {https://doi.org/10.18653/v1/2022.acl-long.62} {Language-agnostic {BERT} sentence embedding}.
\newblock In \emph{Proceedings of the 60th Annual Meeting of the Association for Computational Linguistics (Volume 1: Long Papers)}, pages 878--891, Dublin, Ireland. Association for Computational Linguistics.

\bibitem[{Fuchs et~al.(2022)Fuchs, Ben-shaul, and Mandelbrod}]{fuchs2022}
Gilad Fuchs, Ido Ben-shaul, and Matan Mandelbrod. 2022.
\newblock \href {https://doi.org/10.18653/v1/2022.emnlp-industry.27} {Is it out yet? automatic future product releases extraction from web data}.
\newblock In \emph{Proceedings of the 2022 Conference on Empirical Methods in Natural Language Processing: Industry Track}, pages 263--271, Abu Dhabi, UAE. Association for Computational Linguistics.

\bibitem[{Gong et~al.(2023)Gong, Zhou, Wang, Chen, Song, Cao, Xian, and Zhu}]{gong2023}
Shansan Gong, Zelin Zhou, Shuo Wang, Fengjiao Chen, Xiujie Song, Xuezhi Cao, Yunsen Xian, and Kenny Zhu. 2023.
\newblock \href {https://doi.org/10.18653/v1/2023.acl-industry.46} {Transferable and efficient: Unifying dynamic multi-domain product categorization}.
\newblock In \emph{Proceedings of the 61st Annual Meeting of the Association for Computational Linguistics (Volume 5: Industry Track)}, pages 476--486, Toronto, Canada. Association for Computational Linguistics.

\bibitem[{Hu et~al.(2023)Hu, Sheng, Cao, Zhu, Wang, Wang, and Jin}]{hu2023}
Beizhe Hu, Qiang Sheng, Juan Cao, Yongchun Zhu, Danding Wang, Zhengjia Wang, and Zhiwei Jin. 2023.
\newblock \href {https://doi.org/10.18653/v1/2023.acl-industry.13} {Learn over past, evolve for future: Forecasting temporal trends for fake news detection}.
\newblock In \emph{Proceedings of the 61st Annual Meeting of the Association for Computational Linguistics (Volume 5: Industry Track)}, pages 116--125, Toronto, Canada. Association for Computational Linguistics.

\bibitem[{Keung et~al.(2020)Keung, Lu, Szarvas, and Smith}]{keung-etal-2020-multilingual}
Phillip Keung, Yichao Lu, Gy{\"o}rgy Szarvas, and Noah~A. Smith. 2020.
\newblock \href {https://doi.org/10.18653/v1/2020.emnlp-main.369} {The multilingual {A}mazon reviews corpus}.
\newblock In \emph{Proceedings of the 2020 Conference on Empirical Methods in Natural Language Processing (EMNLP)}, pages 4563--4568, Online. Association for Computational Linguistics.

\bibitem[{Mamani-Coaquira and Villanueva(2024)}]{yonatanmam2024}
Y.~Mamani-Coaquira and E.~Villanueva. 2024.
\newblock \href {https://doi.org/10.1109/ACCESS.2024.3513321} {A review on text sentiment analysis with machine learning and deep learning techniques}.
\newblock \emph{IEEE Access}, 12:193115--193130.

\bibitem[{Mroczkowski et~al.(2021)Mroczkowski, Rybak, Wr{\'o}blewska, and Gawlik}]{mroczkowski2021}
Robert Mroczkowski, Piotr Rybak, Alina Wr{\'o}blewska, and Ireneusz Gawlik. 2021.
\newblock \href {https://www.aclweb.org/anthology/2021.bsnlp-1.1} {{H}er{BERT}: Efficiently pretrained transformer-based language model for {P}olish}.
\newblock In \emph{Proceedings of the 8th Workshop on Balto-Slavic Natural Language Processing}, pages 1--10, Kiyv, Ukraine. Association for Computational Linguistics.

\bibitem[{Nayak and Garera(2022)}]{nayak2022}
Ravindra Nayak and Nikesh Garera. 2022.
\newblock \href {https://doi.org/10.18653/v1/2022.emnlp-industry.55} {Deploying unified {BERT} moderation model for {E}-commerce reviews}.
\newblock In \emph{Proceedings of the 2022 Conference on Empirical Methods in Natural Language Processing: Industry Track}, pages 540--547, Abu Dhabi, UAE. Association for Computational Linguistics.

\bibitem[{Obadinma et~al.(2022)Obadinma, Khan~Khattak, Wang, Sidhorn, Lau, Robertson, Niu, Au, Munim, and Kalaiselvi~Bhaskar}]{obadinma2022}
Stephen Obadinma, Faiza Khan~Khattak, Shirley Wang, Tania Sidhorn, Elaine Lau, Sean Robertson, Jingcheng Niu, Winnie Au, Alif Munim, and Karthik~Raja Kalaiselvi~Bhaskar. 2022.
\newblock \href {https://doi.org/10.18653/v1/2022.emnlp-industry.44} {Bringing the state-of-the-art to customers: A neural agent assistant framework for customer service support}.
\newblock In \emph{Proceedings of the 2022 Conference on Empirical Methods in Natural Language Processing: Industry Track}, pages 440--450, Abu Dhabi, UAE. Association for Computational Linguistics.

\bibitem[{OpenAI et~al.(2024)OpenAI, Achiam, Adler, Agarwal, Ahmad, Akkaya, Aleman, Almeida, Altenschmidt, Altman, Anadkat, Avila, Babuschkin, Balaji, Balcom, Baltescu, Bao, Bavarian, Belgum, Bello, Berdine, Bernadett-Shapiro, Berner, Bogdonoff, Boiko, Boyd, Brakman, Brockman, Brooks, Brundage, Button, Cai, Campbell, Cann, Carey, Carlson, Carmichael, Chan, Chang, Chantzis, Chen, Chen, Chen, Chen, Chen, Chess, Cho, Chu, Chung, Cummings, Currier, Dai, Decareaux, Degry, Deutsch, Deville, Dhar, Dohan, Dowling, Dunning, Ecoffet, Eleti, Eloundou, Farhi, Fedus, Felix, Fishman, Forte, Fulford, Gao, Georges, Gibson, Goel, Gogineni, Goh, Gontijo-Lopes, Gordon, Grafstein, Gray, Greene, Gross, Gu, Guo, Hallacy, Han, Harris, He, Heaton, Heidecke, Hesse, Hickey, Hickey, Hoeschele, Houghton, Hsu, Hu, Hu, Huizinga, Jain, Jain, Jang, Jiang, Jiang, Jin, Jin, Jomoto, Jonn, Jun, Kaftan, Łukasz Kaiser, Kamali, Kanitscheider, Keskar, Khan, Kilpatrick, Kim, Kim, Kim, Kirchner, Kiros, Knight, Kokotajlo, Łukasz Kondraciuk,
  Kondrich, Konstantinidis, Kosic, Krueger, Kuo, Lampe, Lan, Lee, Leike, Leung, Levy, Li, Lim, Lin, Lin, Litwin, Lopez, Lowe, Lue, Makanju, Malfacini, Manning, Markov, Markovski, Martin, Mayer, Mayne, McGrew, McKinney, McLeavey, McMillan, McNeil, Medina, Mehta, Menick, Metz, Mishchenko, Mishkin, Monaco, Morikawa, Mossing, Mu, Murati, Murk, Mély, Nair, Nakano, Nayak, Neelakantan, Ngo, Noh, Ouyang, O'Keefe, Pachocki, Paino, Palermo, Pantuliano, Parascandolo, Parish, Parparita, Passos, Pavlov, Peng, Perelman, de~Avila Belbute~Peres, Petrov, de~Oliveira~Pinto, Michael, Pokorny, Pokrass, Pong, Powell, Power, Power, Proehl, Puri, Radford, Rae, Ramesh, Raymond, Real, Rimbach, Ross, Rotsted, Roussez, Ryder, Saltarelli, Sanders, Santurkar, Sastry, Schmidt, Schnurr, Schulman, Selsam, Sheppard, Sherbakov, Shieh, Shoker, Shyam, Sidor, Sigler, Simens, Sitkin, Slama, Sohl, Sokolowsky, Song, Staudacher, Such, Summers, Sutskever, Tang, Tezak, Thompson, Tillet, Tootoonchian, Tseng, Tuggle, Turley, Tworek, Uribe, Vallone,
  Vijayvergiya, Voss, Wainwright, Wang, Wang, Wang, Ward, Wei, Weinmann, Welihinda, Welinder, Weng, Weng, Wiethoff, Willner, Winter, Wolrich, Wong, Workman, Wu, Wu, Wu, Xiao, Xu, Yoo, Yu, Yuan, Zaremba, Zellers, Zhang, Zhang, Zhao, Zheng, Zhuang, Zhuk, and Zoph}]{openai2024}
OpenAI, Josh Achiam, Steven Adler, Sandhini Agarwal, Lama Ahmad, Ilge Akkaya, Florencia~Leoni Aleman, Diogo Almeida, Janko Altenschmidt, Sam Altman, Shyamal Anadkat, Red Avila, Igor Babuschkin, Suchir Balaji, Valerie Balcom, Paul Baltescu, Haiming Bao, Mohammad Bavarian, Jeff Belgum, Irwan Bello, Jake Berdine, Gabriel Bernadett-Shapiro, Christopher Berner, Lenny Bogdonoff, Oleg Boiko, Madelaine Boyd, Anna-Luisa Brakman, Greg Brockman, Tim Brooks, Miles Brundage, Kevin Button, Trevor Cai, Rosie Campbell, Andrew Cann, Brittany Carey, Chelsea Carlson, Rory Carmichael, Brooke Chan, Che Chang, Fotis Chantzis, Derek Chen, Sully Chen, Ruby Chen, Jason Chen, Mark Chen, Ben Chess, Chester Cho, Casey Chu, Hyung~Won Chung, Dave Cummings, Jeremiah Currier, Yunxing Dai, Cory Decareaux, Thomas Degry, Noah Deutsch, Damien Deville, Arka Dhar, David Dohan, Steve Dowling, Sheila Dunning, Adrien Ecoffet, Atty Eleti, Tyna Eloundou, David Farhi, Liam Fedus, Niko Felix, Simón~Posada Fishman, Juston Forte, Isabella Fulford, Leo
  Gao, Elie Georges, Christian Gibson, Vik Goel, Tarun Gogineni, Gabriel Goh, Rapha Gontijo-Lopes, Jonathan Gordon, Morgan Grafstein, Scott Gray, Ryan Greene, Joshua Gross, Shixiang~Shane Gu, Yufei Guo, Chris Hallacy, Jesse Han, Jeff Harris, Yuchen He, Mike Heaton, Johannes Heidecke, Chris Hesse, Alan Hickey, Wade Hickey, Peter Hoeschele, Brandon Houghton, Kenny Hsu, Shengli Hu, Xin Hu, Joost Huizinga, Shantanu Jain, Shawn Jain, Joanne Jang, Angela Jiang, Roger Jiang, Haozhun Jin, Denny Jin, Shino Jomoto, Billie Jonn, Heewoo Jun, Tomer Kaftan, Łukasz Kaiser, Ali Kamali, Ingmar Kanitscheider, Nitish~Shirish Keskar, Tabarak Khan, Logan Kilpatrick, Jong~Wook Kim, Christina Kim, Yongjik Kim, Jan~Hendrik Kirchner, Jamie Kiros, Matt Knight, Daniel Kokotajlo, Łukasz Kondraciuk, Andrew Kondrich, Aris Konstantinidis, Kyle Kosic, Gretchen Krueger, Vishal Kuo, Michael Lampe, Ikai Lan, Teddy Lee, Jan Leike, Jade Leung, Daniel Levy, Chak~Ming Li, Rachel Lim, Molly Lin, Stephanie Lin, Mateusz Litwin, Theresa Lopez, Ryan
  Lowe, Patricia Lue, Anna Makanju, Kim Malfacini, Sam Manning, Todor Markov, Yaniv Markovski, Bianca Martin, Katie Mayer, Andrew Mayne, Bob McGrew, Scott~Mayer McKinney, Christine McLeavey, Paul McMillan, Jake McNeil, David Medina, Aalok Mehta, Jacob Menick, Luke Metz, Andrey Mishchenko, Pamela Mishkin, Vinnie Monaco, Evan Morikawa, Daniel Mossing, Tong Mu, Mira Murati, Oleg Murk, David Mély, Ashvin Nair, Reiichiro Nakano, Rajeev Nayak, Arvind Neelakantan, Richard Ngo, Hyeonwoo Noh, Long Ouyang, Cullen O'Keefe, Jakub Pachocki, Alex Paino, Joe Palermo, Ashley Pantuliano, Giambattista Parascandolo, Joel Parish, Emy Parparita, Alex Passos, Mikhail Pavlov, Andrew Peng, Adam Perelman, Filipe de~Avila Belbute~Peres, Michael Petrov, Henrique~Ponde de~Oliveira~Pinto, Michael, Pokorny, Michelle Pokrass, Vitchyr~H. Pong, Tolly Powell, Alethea Power, Boris Power, Elizabeth Proehl, Raul Puri, Alec Radford, Jack Rae, Aditya Ramesh, Cameron Raymond, Francis Real, Kendra Rimbach, Carl Ross, Bob Rotsted, Henri Roussez,
  Nick Ryder, Mario Saltarelli, Ted Sanders, Shibani Santurkar, Girish Sastry, Heather Schmidt, David Schnurr, John Schulman, Daniel Selsam, Kyla Sheppard, Toki Sherbakov, Jessica Shieh, Sarah Shoker, Pranav Shyam, Szymon Sidor, Eric Sigler, Maddie Simens, Jordan Sitkin, Katarina Slama, Ian Sohl, Benjamin Sokolowsky, Yang Song, Natalie Staudacher, Felipe~Petroski Such, Natalie Summers, Ilya Sutskever, Jie Tang, Nikolas Tezak, Madeleine~B. Thompson, Phil Tillet, Amin Tootoonchian, Elizabeth Tseng, Preston Tuggle, Nick Turley, Jerry Tworek, Juan Felipe~Cerón Uribe, Andrea Vallone, Arun Vijayvergiya, Chelsea Voss, Carroll Wainwright, Justin~Jay Wang, Alvin Wang, Ben Wang, Jonathan Ward, Jason Wei, CJ~Weinmann, Akila Welihinda, Peter Welinder, Jiayi Weng, Lilian Weng, Matt Wiethoff, Dave Willner, Clemens Winter, Samuel Wolrich, Hannah Wong, Lauren Workman, Sherwin Wu, Jeff Wu, Michael Wu, Kai Xiao, Tao Xu, Sarah Yoo, Kevin Yu, Qiming Yuan, Wojciech Zaremba, Rowan Zellers, Chong Zhang, Marvin Zhang, Shengjia
  Zhao, Tianhao Zheng, Juntang Zhuang, William Zhuk, and Barret Zoph. 2024.
\newblock \href {https://arxiv.org/abs/2303.08774} {Gpt-4 technical report}.
\newblock \emph{Preprint}, arXiv:2303.08774.

\bibitem[{Parikh et~al.(2023)Parikh, Tiwari, Tumbade, and Vohra}]{parikh2023}
Soham Parikh, Mitul Tiwari, Prashil Tumbade, and Quaizar Vohra. 2023.
\newblock \href {https://doi.org/10.18653/v1/2023.acl-industry.71} {Exploring zero and few-shot techniques for intent classification}.
\newblock In \emph{Proceedings of the 61st Annual Meeting of the Association for Computational Linguistics (Volume 5: Industry Track)}, pages 744--751, Toronto, Canada. Association for Computational Linguistics.

\bibitem[{Poświata et~al.(2024)Poświata, Dadas, and Perełkiewicz}]{pl-mteb}
Rafał Poświata, Sławomir Dadas, and Michał Perełkiewicz. 2024.
\newblock \href {https://arxiv.org/abs/2405.10138} {{PL-MTEB: Polish Massive Text Embedding Benchmark}}.
\newblock \emph{Preprint}, arXiv:2405.10138.

\bibitem[{Reimers and Gurevych(2019)}]{sentence-bert}
Nils Reimers and Iryna Gurevych. 2019.
\newblock \href {http://arxiv.org/abs/1908.10084} {Sentence-bert: Sentence embeddings using siamese bert-networks}.
\newblock In \emph{Proceedings of the 2019 Conference on Empirical Methods in Natural Language Processing}. Association for Computational Linguistics.

\bibitem[{{Safe Food Advocacy Europe (SAFE)}()}]{2025SafeFoodAdvocacy}
{Safe Food Advocacy Europe (SAFE)}.
\newblock {Dual Food Quality Project}.
\newblock \url{https://www.safefoodadvocacy.eu/projects/dual-food-quality-project/}.
\newblock [Online; accessed 06-March-2025].

\bibitem[{Satjathanakul and Siriborvornratanakul(2024)}]{joniuhansa2024}
J.~Satjathanakul and T.~Siriborvornratanakul. 2024.
\newblock \href {https://doi.org/10.1007/s41870-024-01907-w} {Sentiment analysis in product reviews in thai language}.
\newblock \emph{International Journal of Information Technology (Singapore)}.

\bibitem[{Shen et~al.(2023)Shen, Asai, Byrne, and De~Gispert}]{shen2023}
Xiaoyu Shen, Akari Asai, Bill Byrne, and Adria De~Gispert. 2023.
\newblock \href {https://doi.org/10.18653/v1/2023.acl-industry.12} {x{PQA}: Cross-lingual product question answering in 12 languages}.
\newblock In \emph{Proceedings of the 61st Annual Meeting of the Association for Computational Linguistics (Volume 5: Industry Track)}, pages 103--115, Toronto, Canada. Association for Computational Linguistics.

\bibitem[{Tercan and Meisen(2022)}]{2022tercan}
Hasan Tercan and Tobias Meisen. 2022.
\newblock \href {https://doi.org/10.1007/s10845-022-01963-8} {Machine learning and deep learning based predictive quality in manufacturing: a systematic review}.

\bibitem[{{The European Consumer Organisation (BEUC)}(2018)}]{2018BEUC}
{The European Consumer Organisation (BEUC)}. 2018.
\newblock { Dual product quality across Europe: state-of-play and the way forward}.
\newblock \url{https://www.beuc.eu/sites/default/files/publications/beuc-x-2018-031_beuc_position_paper_on_dual_quality.pdf}.
\newblock [Online; accessed 06-March-2025].

\bibitem[{Tunstall et~al.(2022)Tunstall, Reimers, Jo, Bates, Korat, Wasserblat, and Pereg}]{setfit}
Lewis Tunstall, Nils Reimers, Unso Eun~Seo Jo, Luke Bates, Daniel Korat, Moshe Wasserblat, and Oren Pereg. 2022.
\newblock \href {https://doi.org/10.48550/ARXIV.2209.11055} {Efficient few-shot learning without prompts}.
\newblock \emph{arXiv preprint}.

\bibitem[{Veselovská(2022)}]{2021lenka}
Lenka Veselovská. 2022.
\newblock \href {https://doi.org/10.1080/14783363.2021.1940929} {Dual quality of products in europe: a serious problem or a marketing opportunity?}
\newblock \emph{Total Quality Management and Business Excellence}, 33.

\bibitem[{Wang et~al.(2024)Wang, Yang, Huang, Yang, Majumder, and Wei}]{wang2024multilingual}
Liang Wang, Nan Yang, Xiaolong Huang, Linjun Yang, Rangan Majumder, and Furu Wei. 2024.
\newblock Multilingual e5 text embeddings: A technical report.
\newblock \emph{arXiv preprint arXiv:2402.05672}.

\bibitem[{Wang et~al.(2023)Wang, Chen, Zhu, Lee, and Gao}]{wang2023}
Tianqi Wang, Lei Chen, Xiaodan Zhu, Younghun Lee, and Jing Gao. 2023.
\newblock \href {https://doi.org/10.18653/v1/2023.acl-industry.55} {Weighted contrastive learning with false negative control to help long-tailed product classification}.
\newblock In \emph{Proceedings of the 61st Annual Meeting of the Association for Computational Linguistics (Volume 5: Industry Track)}, pages 574--580, Toronto, Canada. Association for Computational Linguistics.

\bibitem[{Zhang et~al.(2024)Zhang, Zhang, Long, Xie, Dai, Tang, Lin, Yang, Xie, Huang et~al.}]{mgte}
Xin Zhang, Yanzhao Zhang, Dingkun Long, Wen Xie, Ziqi Dai, Jialong Tang, Huan Lin, Baosong Yang, Pengjun Xie, Fei Huang, et~al. 2024.
\newblock mgte: Generalized long-context text representation and reranking models for multilingual text retrieval.
\newblock In \emph{Proceedings of the 2024 Conference on Empirical Methods in Natural Language Processing: Industry Track}, pages 1393--1412.

\bibitem[{Závadský and Hiadlovský(2020)}]{2020janzav}
Ján Závadský and Vladimír Hiadlovský. 2020.
\newblock \href {https://doi.org/10.21003/EA.V185-07} {Economic problems of dual quality of everyday consumer goods}.
\newblock \emph{Economic Annals-XXI}, 185.

\end{thebibliography}

\appendix
\label{sec:appendix}

\section{Dual Qulity Regulations}
\label{appx:dq_regulations}
The regulatory response to dual quality has evolved significantly within the European Union. The European Commission's 2017 guidelines clarified that while product differentiation is not inherently illegal, misleading consumers violates EU consumer protection laws~\citep{2017EPRS, 2019EPRS}. The Commission's Joint Research Centre (JRC) introduced a harmonized testing methodology to assess product composition variations~\citep{2018ECNewDeal, 2023ECJRC} systematically. Additionally, the Omnibus Directive amended Directive 2005/29/EC, classifying dual quality marketing as misleading when substantial differences exist without a legitimate justification~\citep{2025Chambers}. These measures aim to enhance market transparency and prevent unfair commercial practices. However, challenges remain in enforcement and uniform interpretation across Member States~\citep{2025EUMonitor}. Recent research shows that while the prevalence of dual quality food products declined from 31\% in 2018 to 24\% in 2021, concerns persist regarding non-food items, as similar discrepancies have been identified in household and personal care products~\citep{2023ECJRC}.

Furthermore, consumer advocacy organizations such as BEUC argue that enforcement mechanisms must be strengthened to ensure compliance across all product categories~\citep{2018BEUC}. The SAFE initiative also supports enhanced consumer education and reporting mechanisms to empower individuals to identify and challenge dual quality practices~\citep{2025SafeFoodAdvocacy}. These ongoing legal and regulatory efforts underscore the EU's commitment to fair competition and consumer protection, yet continued vigilance and adaptation of enforcement strategies remain necessary.

\section{DQ Dataset Details}

\subsection{Annotation Process Details}

We established a structured data labelling policy to annotate the data, i.e., assign each opinion or review to its appropriate category. This policy provides clear classification criteria for opinions categorized as \textit{dual quality}, \textit{other problems}, or \textit{standard} (see Table~\ref{tab:annotation_guidelines} for detailed definitions). The annotation process followed predefined guidelines to ensure consistency and reliability, and where necessary, ambiguous cases were resolved through annotators' review.

Examples of labeled reviews from the DQ database, annotated according to the established data annotation protocol and accompanied by annotator comments, are presented in Table~\ref{tab:examples}.

\begin{table}[!h]
\centering
\scriptsize 
\begin{tabular}{p{0.2\columnwidth}p{0.7\columnwidth}}
\hline
\bf Label & \bf Description \\
\hline
dual quality & The review contains information about the fact that the customer bought the same product in two countries and noticed a difference in quality, performance, composition, etc. It is not necessary to give the exact names of the countries, phrases such as “abroad” or “in our country” are sufficient.  The customer is comparing two same products or groups of products. Indicating a difference in price, availability or using a general statement such as “there are differences between products purchased in France and Poland” are \textbf{NOT} classified as dual quality, but as standard review. \\
\hline
other problems & The review does not identify the problem of dual quality, but provides information about other problems, among which we can distinguish:

\textbf{--} differences in products due to a different place of purchase (same market), place of packaging or batch received,

\textbf{--} problems with the product itself that require deeper analysis e.g., deterioration over time,

\textbf{--} practices that are illegal and/or violate customer rights e.g., the product is probably counterfeit, suspected fraud, misleading the customer, no instructions in the required language, no expiration date, etc..  \\
\hline
standard & A standard product review in which the comments described are about the product itself and do not indicate problems addressed by the labels “dual quality” or “other problems”. \\
\hline
\end{tabular}
\caption{Annotation Guidelines.}
\label{tab:annotation_guidelines}
\end{table}

\begin{figure*}[h]
\centering
\includegraphics[scale=0.44]{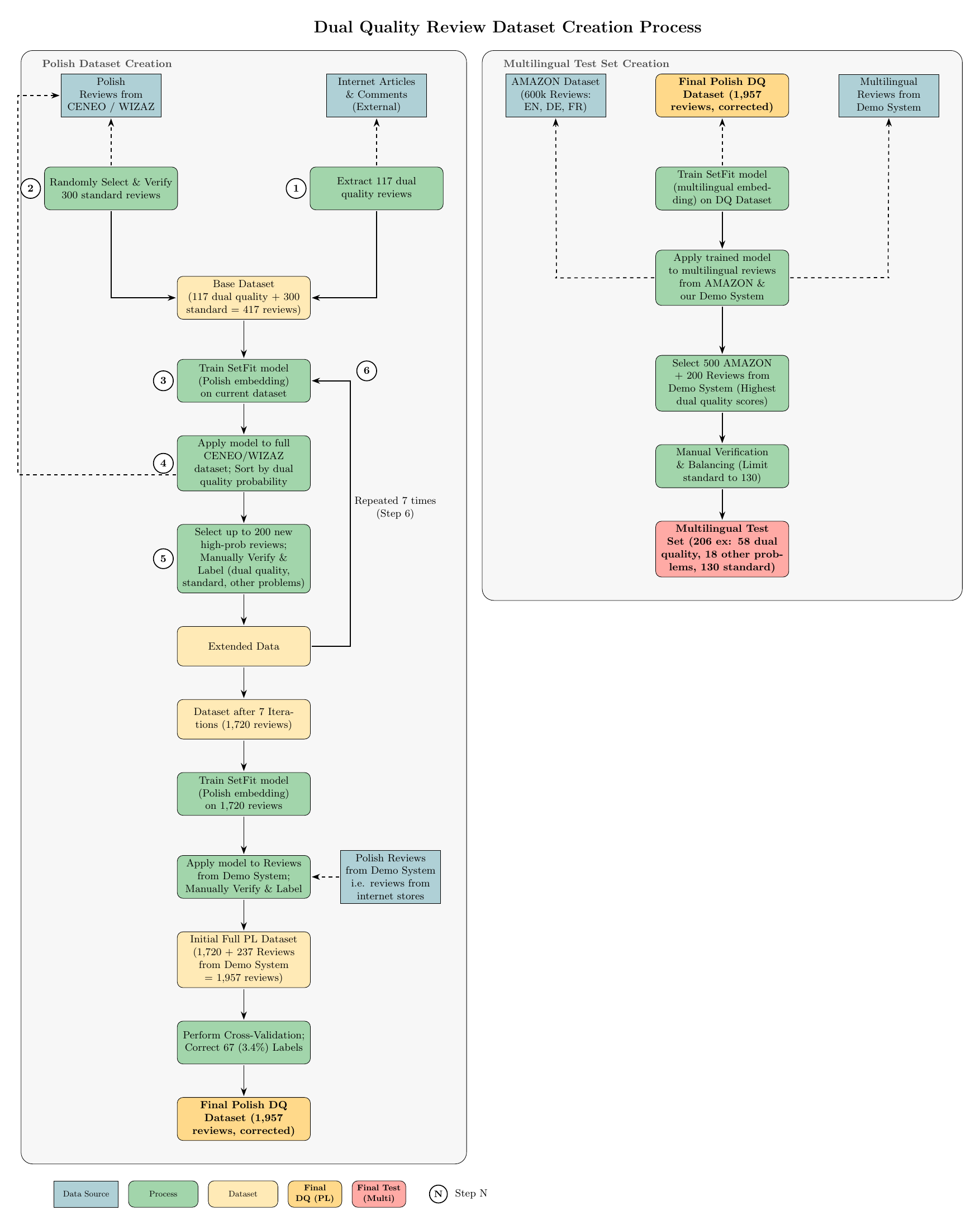}
\caption{Diagram showing the process of preparing DQ and multilingual datasets.}
\label{fig:dataset-creation}
\end{figure*}

\begin{table*}[!h]
\centering
\scriptsize 
\begin{tabular}{p{0.6\columnwidth} | p{0.6\columnwidth} | p{0.3\columnwidth} | p{0.3\columnwidth} }
\hline
 \bf Original review text & \bf Translated review text & \bf Label & \bf Additional Comment \\
\hline
Fantastyczny zapach i produkt z chemii niemieckiej, więc o wiele bardziej intensywny niż te, produkowane na polski rynek. & Fantastic fragrance and a product of German chemistry, so much more intense than those made for the Polish market. & dual quality & - \\
\hline
Jedna z moich ulubionych kaw, zwłaszcza ta w wersji z Niemiec. O wiele bardziej aromatyczna niż proponowana na rynek Polski & One of my favorite coffees, especially the version from Germany. Much more aromatic than the one offered on the Polish market. & dual quality & - \\
\hline
poprzedni model Beko kupiony 9 lat temu był lepszy & The previous Beko model bought 9 years ago was better. & other problems & deterioration in quality over time \\
\hline
Tester w drogerii(w centrum handlowym) był dużo bardziej trwały i intensywniejszy niż ten kupiony przez internet. Zastanawiające. & The tester in the drugstore (at the shopping mall) was much more long-lasting and intense than the one purchased online. Intriguing. & other problems & difference depending on the place of purchase (same market)\\
\hline
Maska spełnia swoje zadanie. Rewelacyjnie pachnie. & The mask does its job. It smells amazing. & stan\-dard & - \\
\hline
soczewki produkowane poza Europą mają kiepską jakość & Lenses produced outside Europe are of poor quality. & stan\-dard & general statement \\
\hline
\end{tabular}
\caption{A list of samples from DQ dataset. The original text of the review was translated
into English using GPT-4o.}
\label{tab:examples}
\end{table*}

\subsection{Other Problems Identified in Products or Services}

\begin{figure*}[!ht]
\centering
\includegraphics[width=0.9\textwidth,keepaspectratio]{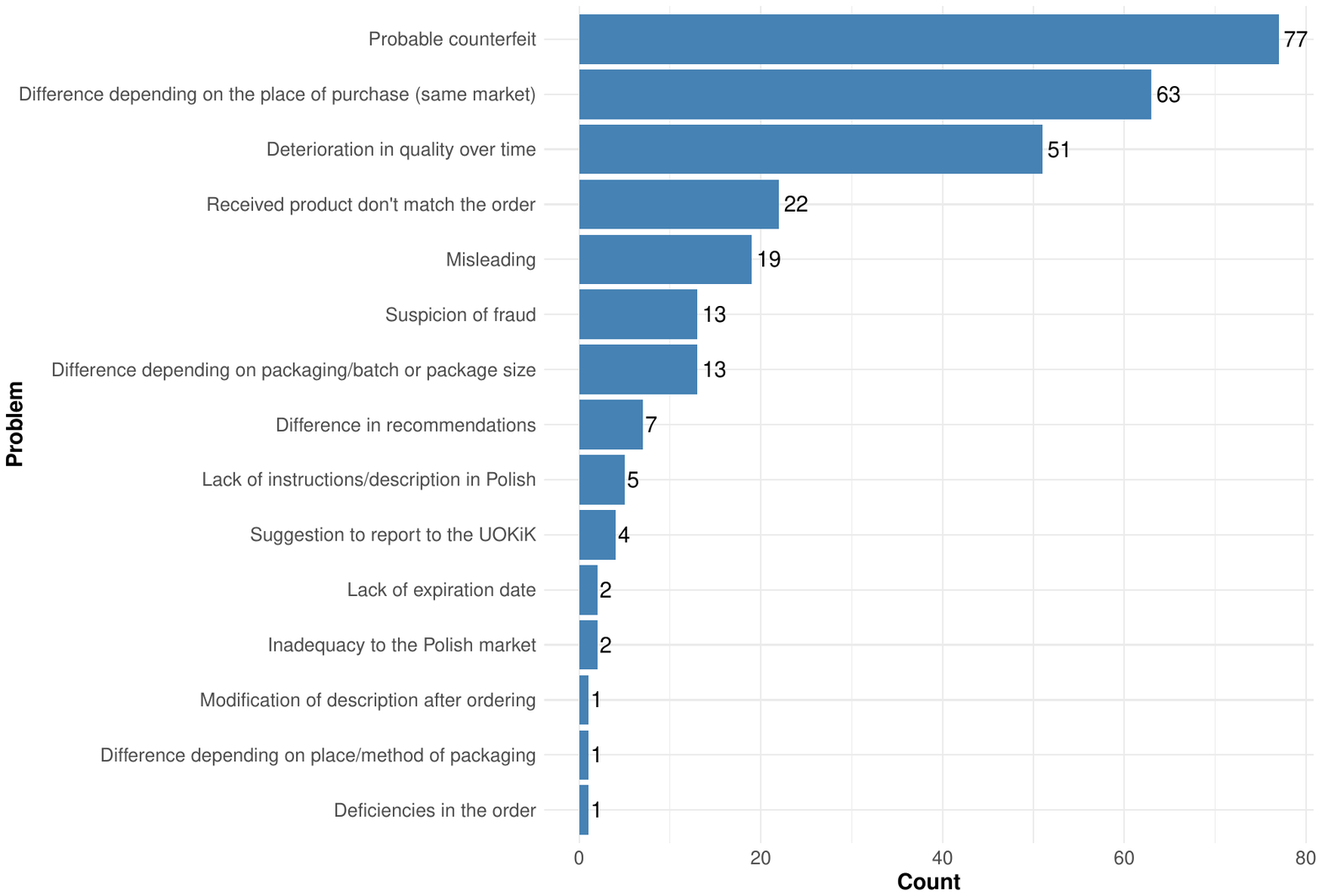}
\caption{Taxonomy of different product or service issues recognized in reviews.}
\label{fig:dataset_preparation}
\end{figure*}

When labeling the data, annotators identified opinions explicitly reflecting dual quality issues and comments pointing to specific problems related to services or products. These additional insights enabled deeper exploration and facilitated the creation of a comprehensive taxonomy of consumer issues. Figure~\ref{fig:dataset_preparation} demonstrates that more than half of the reported problems concern probable counterfeit products, differences dependent on the place of purchase within the same market, quality deterioration over time, mismatches between received products and orders, misleading information, suspicions of fraud, and variations related to packaging, batch, or package size. Recognizing and categorizing these issues may be crucial for targeted interventions and regulatory measures to strengthen consumer trust and improve market standards beyond dual quality considerations alone.

\begin{figure*}[!ht]
\includegraphics[scale=0.455]{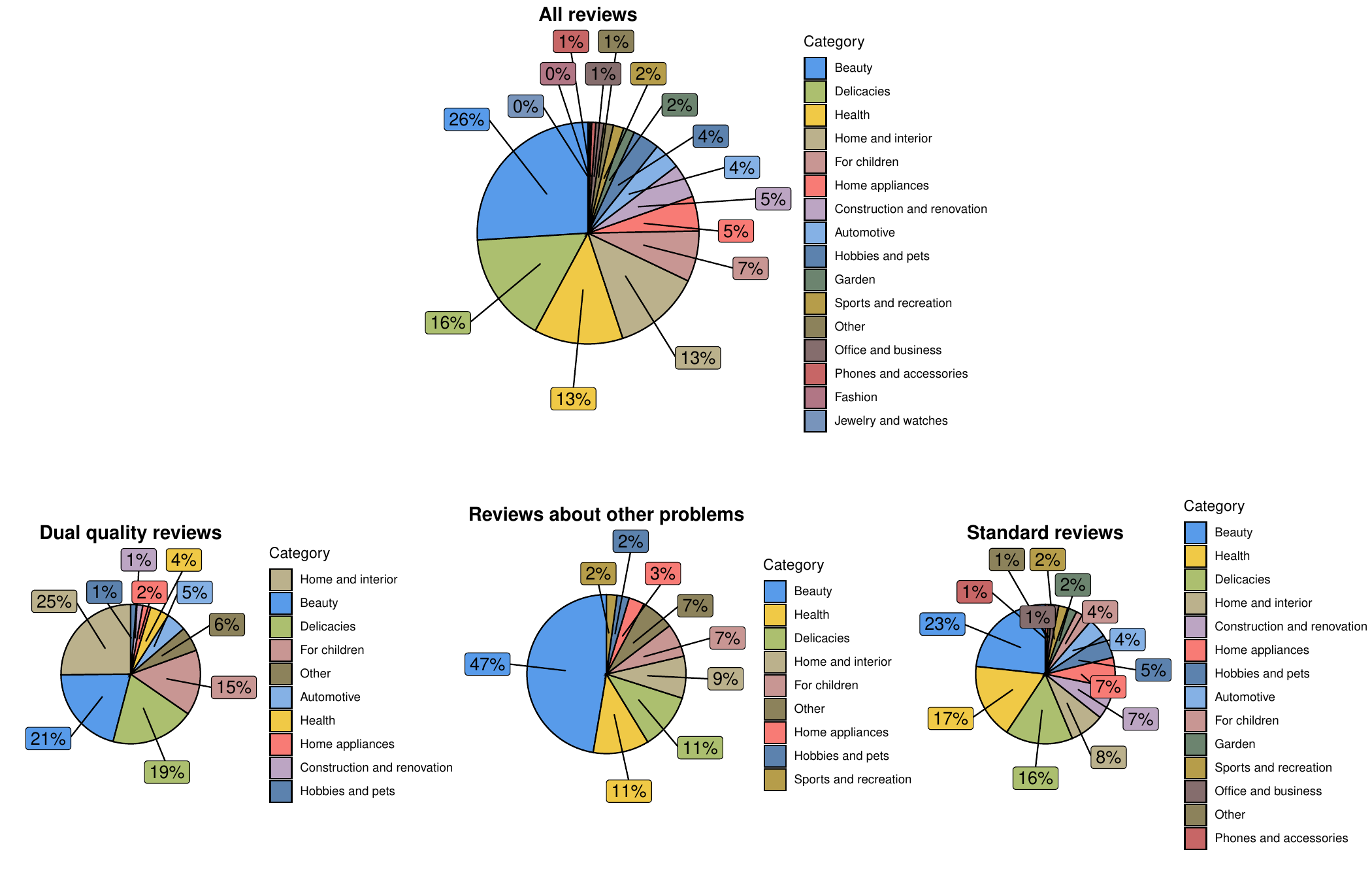}
\caption{Charts illustrating (1) all product reviews categorized by product type (top) and (2) the distribution of product categories across various types of reviews (bottom).}
\label{fig:product_categories}
\end{figure*}

\section{Experiments Details}

\paragraph{Baseline} For the baseline model, the text was first lemmatized. Then the following phrases were searched: \texttt{anglia, angielski, szkocja, szkocki, irlandia, irlandzki, walia, walijski, dania, duński, finlandia, fiński, norwegia, norweski, szwecja, szwedzki, szwajcaria, szwajcarski, estonia, estoński, łotwa, łotewski, litwa, litewski, austria, austryjacki, belgia, belgijski, francja, francuski, niemcy, niemiecki, włochy, włoski, holandia, niderlandzki, holenderski, usa, kanada, kanadyjski, meksyk, meksykański, ukraina, ukraiński, rosja, rosyjski, białoruś, białoruski, polska, polski, czechy, czeski, słowacja, słowacki, węgry, węgierski, rumunia, rumuński, bułgaria, bułgarski, grecja, grecki, hiszpania, hiszpański, brazylia, brazylijski, portugalia, portugalski, australia, australijski, nowa zelandia, maoryjski, gruzja, gruziński, izrael, hebrajski, egipt, arabski, turcja, turecki, chiny, chiński, korea, koreański, japonia, japoński, indie, hinduski}. 

\noindent
If one or more of the above phrases were found, the review was classified as dual quality.

\paragraph{SetFit + sentence transformer} During training, we used the following hyperparameters: learning rate=2e-5 (same for sentence transformer fine-tuning and logistic regression classifier), batch size=8, epochs=1, number of iterations for contrastive=1. We adopted AdamW optimizer.

\paragraph{Transformer-based encoders} During training, we used the following hyperparameters: learning rate=2e-6, batch size=8, epochs=10. We adopted AdamW optimizer.

\paragraph{LLMs}
The models were evaluated using APIs. For the main experiments the temperature was set to 0.1, for robustness verification to guarantee determinism it was reduced to 0.0. The prompts used are shown in Table~\ref{tab:promptspl}.

\begin{table*}
    \centering
    \scriptsize
    \begin{tabular}{ll}
    \toprule
    \bf Name in Paper &  \bf HF Name \\ \midrule 
    LaBSE & sentence-transformers/LaBSE \\
    para-multi-mpnet-base-v2 & sentence-transformers/paraphrase-multilingual-mpnet-base-v2 \\
    para-multi-MiniLM-L12-v2 &  sentence-transformers/paraphrase-multilingual-MiniLM-L12-v2 \\
    multi-e5-small &  intfloat/multilingual-e5-small \\
    multi-e5-base &  intfloat/multilingual-e5-base \\
    multi-e5-large & intfloat/multilingual-e5-large \\
    gte-multi-base & Alibaba-NLP/gte-multilingual-base \\
    st-polish-para-mpnet & sdadas/st-polish-paraphrase-from-mpnet \\
    st-polish-para-distilroberta & sdadas/st-polish-paraphrase-from-distilroberta \\
    mmlw-roberta-base & sdadas/mmlw-roberta-base \\
    mmlw-roberta-large & sdadas/mmlw-roberta-large \\
    mBERT & google-bert/bert-base-multilingual-cased \\
    xlm-roberta-base & FacebookAI/xlm-roberta-base \\
    xlm-roberta-large & FacebookAI/xlm-roberta-large \\
    herbert-base-cased & allegro/herbert-base-cased \\
    herbert-large-cased & allegro/herbert-large-cased \\
    polish-roberta-base-v2 & sdadas/polish-roberta-base-v2 \\
    polish-roberta-large-v2 & sdadas/polish-roberta-large-v2 \\
    deepseek-v3* & deepseek-ai/DeepSeek-V3  \\
    gpt-4o* & - \\
 \bottomrule
    \end{tabular}
    \caption{Model names as referenced in the paper, and corresponding Hugging Face Hub identifiers. An asterisk (*) indicates models accessed via REST APIs: DeepSeek-V3 (\url{https://api-docs.deepseek.com/}) and GPT-4o (\url{https://platform.openai.com/docs/api-reference/introduction}).}
    \label{tab:modelsdesc}
\end{table*}

\begin{table*}[!h]
\scriptsize
\begin{tabularx}{\textwidth}{p{0.15\textwidth}X}
\hline
\textbf{Type} & \textbf{Prompt} \\
\hline
zero-shot & 
Przypisz podaną niżej opinie do jednej z trzech klas: "dual quality", "other problems" lub "standard".\newline
W odpowiedzi podaj jedynie nazwę klasy, bez dodatkowego komentarza.\newline
Treść opinii:\newline
\textit{<review>}
\\
\hline
few-shot & 
Przypisz podaną niżej opinie do jednej z trzech klas: "dual quality", "other problems" lub "standard".\newline\newline
\textbf{Przykłady:}\newline
Kapsułki są lepsze, niż na polski rynek tej samej firmy. -- dual quality\newline
Dobry smak kawy. Kraj pochodzenia Niemcy. Nie jest tak kwaśna jak kupiona w kraju. -- dual quality\newline
Mój ulubiony zapach. Sądzę jednak, że są dużo mniej trwałe niż te, które poprzednim razem kupiłam w sephorze. -- other problems\newline
Proszek może i z Niemiec, ale produkcja Czechy - wprowadzanie klienta w błąd. -- other problems\newline
Niezły preparat. Łagodzi trochę bóle i zmęczenie oczu. Stosuję od czasu do czasu. -- standard\newline
jest ok, nie zauważyłam większej różnicy między "polską" a "niemiecką" wersją -- standard\newline\newline
W odpowiedzi podaj jedynie nazwę klasy, bez dodatkowego komentarza.\newline 
Treść opinii:\newline
\textit{<review>}
\\
\hline
zero-shot+inst. & 
Przypisz podaną niżej opinie do jednej z trzech klas: "dual quality", "other problems" lub "standard".\newline\newline
\textbf{Wytyczne dla każdej z klas:}\newline
\textbf{"dual quality" (podwójna jakość)} -- opinia zawiera informacje o tym, że klient kupił ten sam produkt w dwóch krajach i zauważył różnicę w jakości, wydajności, składzie itp. Nie jest konieczne podawanie dokładnych nazw krajów, wystarczą zwroty takie jak „za granicą” lub „w naszym kraju”. Klient porównuje dwa takie same produkty lub grupy produktów. Wskazanie różnicy w cenie, dostępności lub ogólne stwierdzenie, takie jak „istnieją różnice między produktami zakupionymi we Francji i w Polsce” nie są klasyfikowane jako podwójna jakość.\newline\newline
\textbf{"other problems" (inne problemy)} -- opinia nie wskazuje na problem podwójnej jakości, ale dostarcza informacji o innych problemach, wśród których możemy wyróżnić: różnice w produktach wynikające z innego miejsca zakupu (ten sam rynek), miejsca pakowania lub otrzymanej partii; problemy z samym produktem wymagające głębszej analizy np. pogorszenie jakości z upływem czasu; praktyki niezgodne z prawem i/lub naruszające prawa klienta np. produkt jest prawdopodobnie podrobiony, podejrzenie oszustwa, wprowadzanie klienta w błąd, brak instrukcji w wymaganym języku, brak daty ważności itp.\newline\newline
\textbf{"standard"} -- standardowa opinia o produkcie, w której opisane uwagi dotyczą samego produktu i nie wskazują na problemy omówione przy klasach „podwójna jakość” lub „inne problemy”.\newline\newline
W odpowiedzi podaj jedynie nazwę klasy, bez dodatkowego komentarza.\newline 
Treść opinii:\newline
\textit{<review>}
\\
\hline
few-shot+inst. & 
Przypisz podaną niżej opinie do jednej z trzech klas: "dual quality", "other problems" lub "standard".\newline\newline
\textbf{Wytyczne dla każdej z klas:}\newline
\textbf{"dual quality" (podwójna jakość)} -- opinia zawiera informacje o tym, że klient kupił ten sam produkt w dwóch krajach i zauważył różnicę w jakości, wydajności, składzie itp. Nie jest konieczne podawanie dokładnych nazw krajów, wystarczą zwroty takie jak „za granicą” lub „w naszym kraju”. Klient porównuje dwa takie same produkty lub grupy produktów. Wskazanie różnicy w cenie, dostępności lub ogólne stwierdzenie, takie jak „istnieją różnice między produktami zakupionymi we Francji i w Polsce” nie są klasyfikowane jako podwójna jakość.\newline
\textbf{Przykłady:} "Kapsułki są lepsze, niż na polski rynek tej samej firmy.", "Dobry smak kawy. Kraj pochodzenia Niemcy. Nie jest tak kwaśna jak kupiona w kraju."\newline\newline
\textbf{"other problems" (inne problemy)} -- opinia nie wskazuje na problem podwójnej jakości, ale dostarcza informacji o innych problemach, wśród których możemy wyróżnić: różnice w produktach wynikające z innego miejsca zakupu (ten sam rynek), miejsca pakowania lub otrzymanej partii; problemy z samym produktem wymagające głębszej analizy np. pogorszenie jakości z upływem czasu; praktyki niezgodne z prawem i/lub naruszające prawa klienta np. produkt jest prawdopodobnie podrobiony, podejrzenie oszustwa, wprowadzanie klienta w błąd, brak instrukcji w wymaganym języku, brak daty ważności itp.\newline
\textbf{Przykłady:} "Mój ulubiony zapach. Sądzę jednak, że są dużo mniej trwałe niż te, które poprzednim razem kupiłam w sephorze", "Proszek może i z Niemiec, ale produkcja Czechy - wprowadzanie klienta w błąd."\newline\newline
\textbf{"standard"} -- standardowa opinia o produkcie, w której opisane uwagi dotyczą samego produktu i nie wskazują na problemy omówione przy klasach „podwójna jakość” lub „inne problemy”.\newline
\textbf{Przykłady:} "Niezły preparat. Łagodzi trochę bóle i zmęczenie oczu. Stosuję od czasu do czasu.", "jest ok, nie zauważyłam większej różnicy między "polską" a "niemiecką" wersją"\newline\newline
W odpowiedzi podaj jedynie nazwę klasy, bez dodatkowego komentarza.\newline 
Treść opinii:\newline
\textit{<review>}
\\
\hline
\end{tabularx}
\caption{Prompts used during LLMs evaluation. Bold text and blank lines were added only for readability of the table. For non-Polish speakers, translated prompts available in Table~\ref{tab:promptsen}.}
\label{tab:promptspl}
\end{table*}

\begin{table*}[!h]
\scriptsize
\begin{tabularx}{\textwidth}{p{0.15\textwidth}X}
\hline
\textbf{Type} & \textbf{Prompt} \\
\hline
zero-shot &
Assign the following review to one of three classes: “dual quality”, “other problems” or “standard”.\newline
In your answer, provide only the name of the class, without additional comment.\newline
Review text:\newline
\textit{<review>}
\\
\hline
few-shot &
Assign the following review to one of three classes: “dual quality”, “other problems” or “standard”.\newline\newline
\textbf{Examples:}\newline
The capsules are better than those on the Polish market from the same company. -- dual quality\newline
Good coffee taste. Country of origin: Germany. It is not as acidic as the one bought in the country. -- dual quality\newline
My favorite scent. However, I think it's much less long-lasting than the one I bought at Sephora last time. -- other problems\newline
The powder may be from Germany, but it's made in the Czech Republic - misleading the customer. -- other problems\newline
Decent product. It slightly alleviates eye pain and fatigue. I use it occasionally. -- standard\newline
It's okay, I didn't notice much difference between the "Polish" and "German" version. -- standard\newline\newline
In your answer, provide only the name of the class, without additional comment.\newline
Review text:\newline
\textit{<review>}
\\
\hline
zero-shot+inst. &
Assign the following review to one of three classes: “dual quality”, “other problems” or “standard”.\newline\newline
\textbf{Guidelines for each category:}\newline
\textbf{"dual quality"} -- The review includes information that the customer purchased the same product in two different countries and noticed a difference in quality, performance, composition, etc. It is not necessary to specify the exact names of the countries; phrases like "abroad" or "in our country" are sufficient. The customer compares two identical products or groups of products. Indicating a difference in price, availability, or a general statement such as "there are differences between products purchased in France and Poland" is not classified as dual quality.\newline\newline
\textbf{"other problems"} -- The review does not indicate an issue of dual quality but provides information on other problems, which can include: differences in products resulting from a different place of purchase (same market), place of packaging, or the received batch; problems with the product itself requiring deeper analysis, such as deterioration in quality over time; practices that are illegal and/or violate customer rights, such as the product potentially being counterfeit, suspicion of fraud, misleading the customer, lack of instructions in the required language, lack of an expiration date, etc.
\newline\newline
\textbf{"standard"} -- A standard product review where the comments pertain only to the product itself and do not indicate the problems discussed in the "dual quality" or "other problems" categories.
\newline\newline
In your answer, provide only the name of the class, without additional comment.\newline
Review text:\newline
\textit{<review>}
\\
\hline
few-shot+inst. &
Assign the following review to one of three classes: “dual quality”, “other problems” or “standard”.\newline\newline
\textbf{Guidelines for each category:}\newline
\textbf{"dual quality"} -- The review includes information that the customer purchased the same product in two different countries and noticed a difference in quality, performance, composition, etc. It is not necessary to specify the exact names of the countries; phrases like "abroad" or "in our country" are sufficient. The customer compares two identical products or groups of products. Indicating a difference in price, availability, or a general statement such as "there are differences between products purchased in France and Poland" is not classified as dual quality.\newline
\textbf{Examples:} "The capsules are better than those on the Polish market from the same company.", "Good coffee taste. Country of origin: Germany. It is not as acidic as the one bought in the country."
\newline\newline
\textbf{"other problems"} -- The review does not indicate an issue of dual quality but provides information on other problems, which can include: differences in products resulting from a different place of purchase (same market), place of packaging, or the received batch; problems with the product itself requiring deeper analysis, such as deterioration in quality over time; practices that are illegal and/or violate customer rights, such as the product potentially being counterfeit, suspicion of fraud, misleading the customer, lack of instructions in the required language, lack of an expiration date, etc.
\newline
\textbf{Examples:} "My favorite scent. However, I think it's much less long-lasting than the one I bought at Sephora last time.",
"The powder may be from Germany, but it's made in the Czech Republic - misleading the customer."
\newline\newline
\textbf{"standard"} -- A standard product review where the comments pertain only to the product itself and do not indicate the problems discussed in the "dual quality" or "other problems" categories.\newline
\textbf{Examples:} "Decent product. It slightly alleviates eye pain and fatigue. I use it occasionally.", "It's okay, I didn't notice much difference between the "Polish" and "German" version."
\newline\newline
In your answer, provide only the name of the class, without additional comment.\newline
Review text:\newline
\textit{<review>}
\\
\hline
\end{tabularx}
\caption{Translated prompts from Table~\ref{tab:promptspl} used during LLMs evaluation.}
\label{tab:promptsen}
\end{table*}

\begin{figure*}[h]
\centering
\includegraphics[scale=0.43]{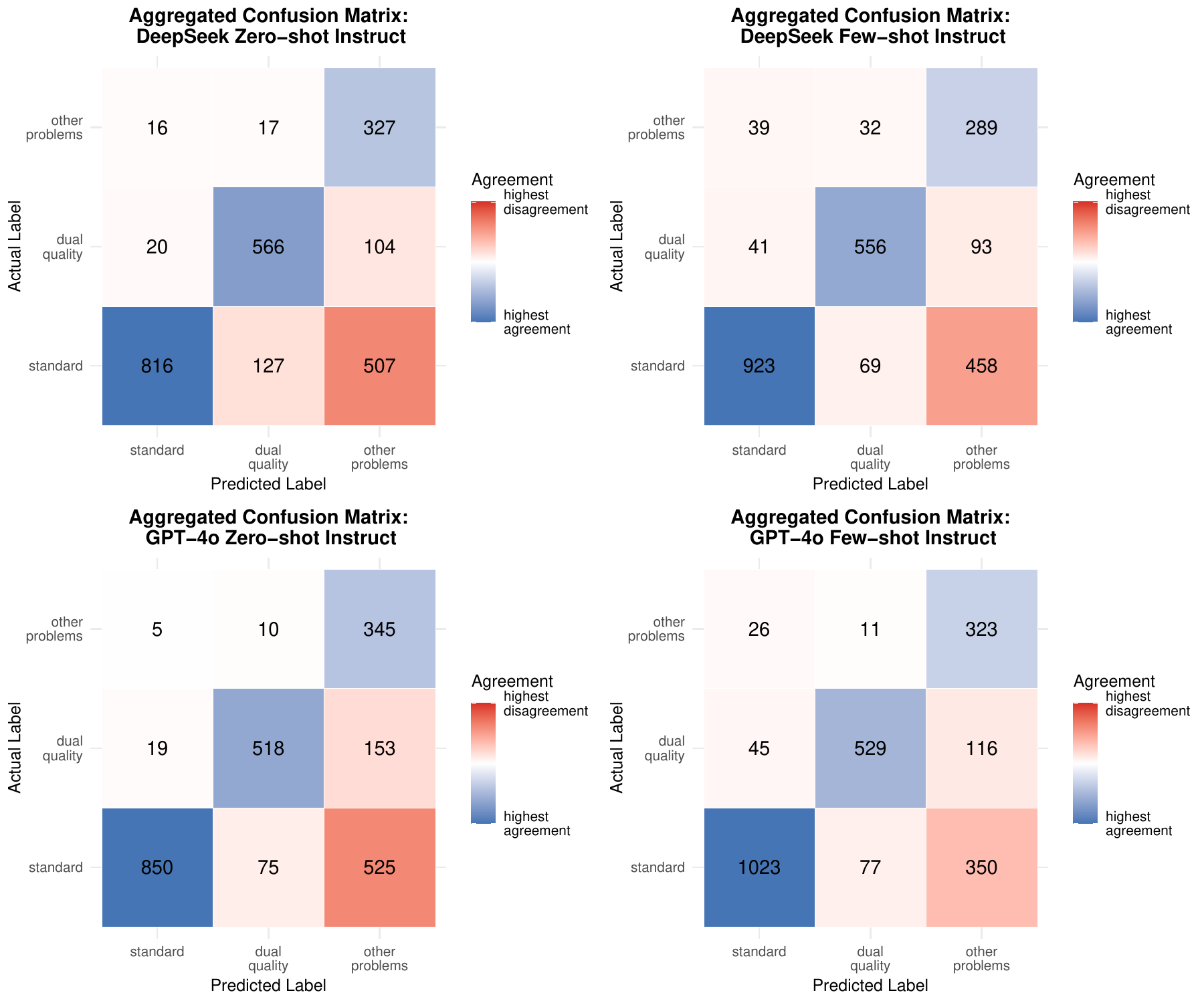}
\caption{Confusion matrices aggregated from five experiments for DeepSeek and GPT-4o models in zero-shot and few-shot instruction-based configurations.}
\label{fig:hm2}
\end{figure*}

\begin{figure*}[h]
\centering
\includegraphics[scale=0.43]{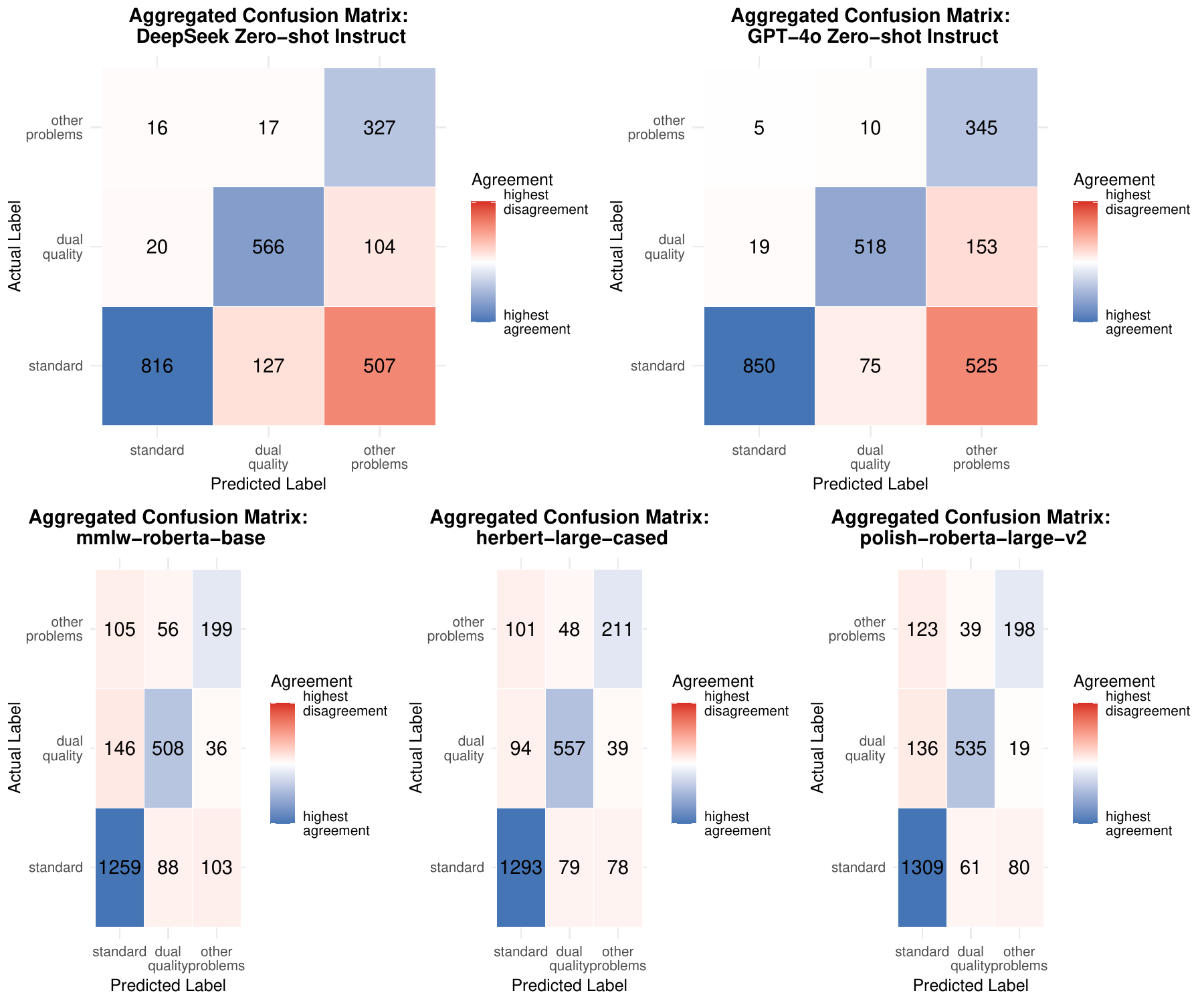}
\caption{Confusion matrices aggregated from five experiments for best performing LLMs and top-performing local models.}
\label{fig:hm3}
\end{figure*}

\begin{table*}[h]
\scriptsize
\centering
\begin{tabular}{llll|llll}
\hline
& \multicolumn{3}{c}{\bf Dual Quality class} & \multicolumn{4}{c}{\bf All classes} \\
 \bf Method  & \bf Precision        & \bf Recall           & \bf F1         & \bf Accuracy        & \bf mPrecision        & \bf mRecall           & \bf mF1         \\
\hline
 \multicolumn{8}{l}{\textbf{SetFit + sentence transformers}} \\
 \hline
  LaBSE                                       & $74.0_{\pm8.7}$                   & $37.9_{\pm12.1}$                  & $49.1_{\pm11.4}$                  & $70.1_{\pm3.6}$                   & $55.5_{\pm3.6}$                   & $47.6_{\pm4.1}$                   & $48.4_{\pm4.8}$                   \\
 para-multi-mpnet-base-v2                    & $69.4_{\pm3.4}$                   & $45.5_{\pm4.4}$                   & $54.8_{\pm3.0}$                   & $67.1_{\pm1.8}$                   & $53.2_{\pm1.4}$                   & $49.6_{\pm0.8}$                   & $50.2_{\pm0.6}$                   \\
 para-multi-MiniLM-L12-v2                    & $69.3_{\pm3.2}$                   & $40.3_{\pm7.8}$                   & $50.7_{\pm7.4}$                   & $62.9_{\pm2.0}$                   & $49.3_{\pm1.9}$                   & $43.2_{\pm2.7}$                   & $44.6_{\pm3.0}$                   \\
 multi-e5-small                              & $74.0_{\pm4.1}$                   & $41.7_{\pm4.8}$                   & $53.2_{\pm4.2}$                   & $72.9_{\pm1.3}$                   & $49.1_{\pm1.5}$                   & $46.2_{\pm1.5}$                   & $45.5_{\pm1.7}$                   \\
 multi-e5-base                               & $78.4_{\pm4.8}$                   & $45.9_{\pm19.1}$                  & $54.6_{\pm19.1}$                  & \color{blue}{$\bf 73.4_{\pm4.1}$} & $54.1_{\pm5.0}$                   & $49.2_{\pm7.1}$                   & $48.6_{\pm9.1}$                   \\
 multi-e5-large                              & \color{red}{$0.0_{\pm0.0}$}                    & \color{red}{$0.0_{\pm0.0}$}                    & \color{red}{$0.0_{\pm0.0}$}                    & $63.1_{\pm0.0}$                   & $21.0_{\pm0.0}$                   & $33.3_{\pm0.0}$                   & $25.8_{\pm0.0}$                   \\
 gte-multi-base                              & $\bf 81.7_{\pm4.9}$               & $\bf 58.0_{\pm4.7}$               & $\bf 67.7_{\pm3.7}$               & $71.6_{\pm3.3}$                   & $\bf 57.2_{\pm2.7}$               & $\bf 52.5_{\pm2.6}$               & $\bf 54.0_{\pm2.7}$               \\
 \hline
 \multicolumn{8}{l}{\textbf{Transformer-based encoders}} \\
 \hline
mBERT                                       & $61.7_{\pm19.5}$                  & $6.6_{\pm4.3}$                    & $11.1_{\pm6.7}$                   & $62.1_{\pm2.8}$                   & $43.8_{\pm6.3}$                   & $34.7_{\pm2.2}$                   & $30.3_{\pm3.3}$                   \\
 xlm-roberta-base                            & $69.5_{\pm2.3}$                   & $\bf 66.9_{\pm6.8}$               & $67.9_{\pm2.9}$                   & $\bf 73.0_{\pm1.0}$               & $55.5_{\pm1.1}$                   & $55.1_{\pm2.1}$                   & $55.0_{\pm1.7}$                   \\
 xlm-roberta-large                           & $\bf 84.8_{\pm3.8}$               & $63.1_{\pm4.8}$                   & \color{blue}{$\bf 72.3_{\pm4.0}$} & $72.6_{\pm2.7}$                   & $\bf 60.1_{\pm2.7}$               & $\bf 56.7_{\pm3.9}$               & \color{blue}{$\bf 57.5_{\pm3.3}$} \\
  \hline
 \multicolumn{8}{l}{\textbf{LLMs}} \\
 \hline
 deepseek-v3 \textsubscript{zero-shot}       & $47.6_{\pm1.9}$                   & $86.2_{\pm2.8}$                   & $61.4_{\pm2.3}$                   & $32.4_{\pm0.9}$                   & $46.9_{\pm1.5}$                   & $39.4_{\pm1.0}$                   & $28.8_{\pm0.9}$                   \\
 deepseek-v3 \textsubscript{few-shot}        & $62.8_{\pm1.4}$                   & $70.7_{\pm1.4}$                   & $\bf 66.5_{\pm0.7}$               & $35.6_{\pm0.6}$                   & $54.3_{\pm0.7}$                   & $46.7_{\pm1.8}$                   & $36.7_{\pm0.7}$                   \\
 deepseek-v3 \textsubscript{zero-shot+inst.} & $85.9_{\pm1.8}$                   & $52.3_{\pm0.8}$                   & $65.0_{\pm0.3}$                   & $49.5_{\pm0.7}$                   & $63.4_{\pm1.3}$                   & \color{blue}{$\bf 58.7_{\pm1.0}$} & $49.1_{\pm0.7}$                   \\
 deepseek-v3 \textsubscript{few-shot+inst.}  & \color{blue}{$\bf 91.9_{\pm4.8}$} & $50.6_{\pm0.8}$                   & $65.2_{\pm1.8}$                   & $44.3_{\pm0.9}$                   & \color{blue}{$\bf 65.6_{\pm2.2}$} & $56.2_{\pm1.2}$                   & $46.1_{\pm1.0}$                   \\
 gpt-4o \textsubscript{zero-shot}            & $38.8_{\pm0.6}$                   & \color{blue}{$\bf 86.8_{\pm2.2}$} & $53.6_{\pm1.0}$                   & $33.3_{\pm0.6}$                   & $47.4_{\pm0.2}$                   & $36.8_{\pm0.7}$                   & $27.0_{\pm0.4}$                   \\
 gpt-4o \textsubscript{few-shot}             & $58.5_{\pm0.8}$                   & $73.6_{\pm0.8}$                   & $65.1_{\pm0.4}$                   & $34.1_{\pm0.6}$                   & $55.8_{\pm0.6}$                   & $48.1_{\pm2.3}$                   & $34.7_{\pm0.7}$                   \\
 gpt-4o \textsubscript{zero-shot+inst.}      & $85.3_{\pm1.3}$                   & $46.6_{\pm0.0}$                   & $60.2_{\pm0.3}$                   & $\bf 52.6_{\pm0.6}$               & $62.3_{\pm0.3}$                   & $57.1_{\pm0.3}$                   & $\bf 49.6_{\pm0.3}$               \\
 gpt-4o \textsubscript{few-shot+inst.}       & $80.2_{\pm1.1}$                   & $46.6_{\pm0.0}$                   & $58.9_{\pm0.3}$                   & $41.6_{\pm0.6}$                   & $61.4_{\pm0.5}$                   & $50.2_{\pm1.0}$                   & $42.7_{\pm0.5}$                   \\
\hline
\end{tabular}
\caption{\label{tab:multi-results-full}
Evaluation results on a multilingual dataset consisting of English, German and French reviews. In red were marked results showing an example of when a multilingual transfer did not work.}
\end{table*}

\end{document}